\documentclass{article}

\usepackage[final]{corl_2019} %

\usepackage{amsmath,amsfonts,bm}

\def\eqref#1{equation~\ref{#1}}
\def\Eqref#1{Equation~\ref{#1}}

\def\1{\bm{1}}

\def\eps{{\epsilon}}

\def\vmu{{\bm{\mu}}}
\def\vtheta{{\bm{\theta}}}
\def\va{{\bm{a}}}

\def\ve{{\bm{e}}}
\def\vf{{\bm{f}}}
\def\vg{{\bm{g}}}

\def\vk{{\bm{k}}}

\def\vp{{\bm{p}}}

\def\vu{{\bm{u}}}

\def\vx{{\bm{x}}}
\def\vy{{\bm{y}}}
\def\vz{{\bm{z}}}

\def\mA{{\bm{A}}}
\def\mB{{\bm{B}}}

\def\mE{{\bm{E}}}
\def\mF{{\bm{F}}}

\def\mI{{\bm{I}}}
\def\mJ{{\bm{J}}}
\def\mK{{\bm{K}}}

\def\mP{{\bm{P}}}
\def\mQ{{\bm{Q}}}
\def\mR{{\bm{R}}}

\def\mU{{\bm{U}}}

\def\mX{{\bm{X}}}

\def\mZ{{\bm{Z}}}

\def\mLambda{{\bm{\Lambda}}}
\def\mSigma{{\bm{\Sigma}}}

\DeclareMathAlphabet{\mathsfit}{\encodingdefault}{\sfdefault}{m}{sl}
\SetMathAlphabet{\mathsfit}{bold}{\encodingdefault}{\sfdefault}{bx}{n}

\usepackage{amsmath}
\usepackage{amssymb}
\usepackage{fixmath}
\newcommand{\mul}[3]{\mathbold{\mu}_{\overleftarrow{#1}_{#3}^{#2}}}
\newcommand{\mur}[3]{\mathbold{\mu}_{\overrightarrow{#1}_{#3}^{#2}}}
\newcommand{\mum}[3]{\mathbold{\mu}_{{#1}_{#3}^{#2}}}

\newcommand{\sigl}[3]{\mathbold{\Sigma}_{\overleftarrow{#1}_{#3}^{#2}}}
\newcommand{\sigr}[3]{\mathbold{\Sigma}_{\overrightarrow{#1}_{#3}^{#2}}}
\newcommand{\sigm}[3]{\mathbold{\Sigma}_{{#1}_{#3}^{#2}}}

\newcommand{\nul}[3]{\mathbold{\nu}_{\overleftarrow{#1}_{#3}^{#2}}}
\newcommand{\nur}[3]{\mathbold{\nu}_{\overrightarrow{#1}_{#3}^{#2}}}

\newcommand{\nua}[3]{\Tilde{\mathbold{\nu}}_{{#1}_{#3}^{#2}}}
\newcommand{\laml}[3]{\mathbold{\Lambda}_{\overleftarrow{#1}_{#3}^{#2}}}
\newcommand{\lamr}[3]{\mathbold{\Lambda}_{\overrightarrow{#1}_{#3}^{#2}}}
\newcommand{\lamm}[3]{\mathbold{\Lambda}_{{#1}_{#3}^{#2}}}
\newcommand{\lama}[3]{\Tilde{\mathbold{\Lambda}}_{{#1}_{#3}^{#2}}}

\newcommand{\mat}[1]{\mathbold{#1}}
\newcommand{\vnu}{\mat{\nu}}

\newcommand{\veta}{\mat{\eta}}
\newcommand{\vxi}{\mat{\xi}}

\newcommand{\mGamma}{\mat{\Gamma}}
\newcommand{\mPsi}{\mat{\Psi}}

\newcommand{\sigEta}[1]{\mat{\Sigma}_{\veta_{#1}}}
\newcommand{\sigXi}{\mat{\Sigma}_{\vxi}}
\newcommand{\lamXi}{\mat{\Lambda}_{\vxi}}
\newcommand{\tran}{^\intercal}
\newcommand{\inv}{^{\text{-}1}}
\usepackage{algorithm2e}
\usepackage{hyperref}
\usepackage{url}
\usepackage{graphicx}
\usepackage{amsmath}
\usepackage{amsfonts}
\usepackage{amssymb}
\usepackage{float}
\usepackage{caption}
\usepackage{tikz}
\usetikzlibrary{arrows}
\usetikzlibrary{fit}
\usepackage{xcolor}
\usepackage{multirow}
\usepackage{subcaption}
\usepackage{wrapfig}
\usepackage{mathtools}
\usepackage{ragged2e}
\usepackage[export]{adjustbox}
\usepackage{xr-hyper}
\usepackage{booktabs}
\usepackage{array}
\usepackage[nice]{nicefrac}
\usepackage{enumitem}
\usepackage{xspace}
\usepackage{natbib}

\usepackage{pgfplots}
\pgfplotsset{compat=newest}
\usepgfplotslibrary{groupplots}
\usepgfplotslibrary{dateplot}

\DeclareMathOperator{\tr}{tr}

\newcommand{\itwoc}{\textsc{i2c}}

\providecommand{\sn}[1]{\ensuremath{\times10^{#1}}}

\newcommand{\kld}[2]{D_{\text{KL}}\!\left(\!\left. \left. #1 \right|\right| #2 \right) }

\title{Stochastic Optimal Control as\\
Approximate Input Inference}

\author{Joe Watson, Hany Abdulsamad, Jan Peters$\dagger$\\
Department of Computer Science, Technische Universit\"at Darmstadt, Germany\\
$\dagger$Robot Learning Group, Max Planck Institute for Intelligent Systems,T\"ubingen, Germany\\
\texttt{\{watson, abdulsamad, peters\}@ias.informatik.tu-darmstadt.de}
}

\newcommand{\altPhrase}[1]{}

\begin{document}
\maketitle

\begin{abstract}
	Optimal control of stochastic nonlinear dynamical systems is a major challenge in the domain of robot learning.
	Given the intractability of the global control problem, state-of-the-art algorithms focus on approximate sequential optimization techniques, that heavily rely on heuristics for regularization in order to achieve stable convergence.
	By building upon the duality between inference and control, we develop the view of Optimal Control as Input Estimation, devising a probabilistic stochastic optimal control formulation that iteratively infers the optimal input distributions by minimizing an upper bound of the control cost.
	Inference is performed through Expectation Maximization and message passing on a probabilistic graphical model of the dynamical system, and time-varying linear Gaussian feedback controllers are extracted from the joint state-action distribution.
	This perspective incorporates uncertainty quantification, effective initialization through priors, and the principled regularization inherent to the Bayesian treatment.
	Moreover, it can be shown that for deterministic linearized systems, our framework derives the maximum entropy linear quadratic optimal control law.
	We provide a complete and detailed derivation of our probabilistic approach and highlight its advantages in comparison to other deterministic and probabilistic solvers.
\end{abstract}


\section{Introduction}
Trajectory Optimization for nonlinear dynamical systems is among the most fundamental paradigms in the field of robotics. It has proven itself to be a cornerstone for both low- and high-level planning techniques \cite{tassa2012synthesis, toussaint2018differentiable}.
A popular tool for devising such planning schemes is Optimal Control \cite{kirk2012optimal, stengel1986stochastic}, which frames the search for the best sequence of inputs into a dynamical system as the optimization of the state-action trajectory.
While Optimal Control has had great success both in theory and application, mainly represented by Sequential Quadratic Programming (SQP) techniques \cite{bryson2018applied}, it is known to struggle with stochastic environments due to its feedforward nature.
Meanwhile, a popular tool for dealing with uncertainty is Bayesian statistics \cite{Bishop:2006:PRM:1162264}, which in part uses the notion of random variables to describe model uncertainty.
The process of determining the characteristics of this uncertainty is known as inference, and this too is often framed as an optimization problem.
Control-as-inference \cite{Attias03planningby, toussaint2006probabilistic, DBLP:conf/aips/KappenGO13, levine2018reinforcement} is a body of research combining these two paradigms, with the proposition that the principled mechanisms of inference will bring the benefits of faster convergence, more principled regularization and the addition of uncertainty quantification \cite{hennig2015probabilistic}.

In this work we present Input Inference for Control (\itwoc{}), a new perspective on control-as-inference.
By moving away from the typical Optimal Control formulation, while preserving the underlying operations, recursive Bayesian inference can be applied to the inputs to manner that optimizes a control objective.
This builds on previous work that performs recursive approximate inference of the state trajectory \cite{toussaint2009robot} and exact input inference for linear systems \cite{hoffmann2017linear}.
Consider the fundamental task of control: to find the sequence of actions that generate a desired trajectory.
From an inference perspective, we would call this problem Input Estimation, where the `desired' observed trajectory in this case is a set of measurements.
In Optimal Control, as the desired trajectory is not expected to be fully achieved, the notion of a cost function is used to describe the desired deviation of the observed trajectory.
Statistically this deviation would be framed as a `disturbance' and described by a probability distribution.
As likelihoods are often the optimization objective of an inference problem, by comparing the likelihood of this formulation to typical control cost functions informs our choice of disturbance noise in order to achieve equivalence.
In this work, we focus on the well-established duality between Gaussian noise and quadratic penalties \cite{Bishop:2006:PRM:1162264}.

By making the linear Gaussian assumption on both our dynamics and observation models, inference can be performed in closed-form using message passing, and we show that this input inference reduces to the Linear Quadratic Regulator (LQR) solution in the deterministic case.
Moreover, the inference is in fact performing the same Discrete Algebraic Ricatti equation (DARE) computation \cite{hoffmann2017linear}.
Additionally, making the inference approximate through local linearizations, we extend the scheme to nonlinear dynamical systems and arrive at a procedure akin to the popular trajectory optimization of Differential Dynamic Programming (DDP) \cite{jacobson1970differential} and variants (e.g. iLQR \cite{li2004iterative}, eLQR\cite{van2016extended}, GPS \cite{DBLP:conf/icml/LevineK13}).
While these methods require explicit regularization, bounds and heuristics to maintain steady convergence, the behaviour of our scheme is governed primarily by the choice of priors, and the regularization only required to account for the log-likelihood approximation.
The use of Bayesian inference also results in self-regularized exploration, as the covariance of each input is a measure of confidence / robustness.
Moreover, by examining the conditional distributions between the resultant posterior state-action distribution, we arrive at (Bayes) optimal time-varying linear (Gaussian) controllers, as in LQR  \cite{hoffmann2017linear}.
We show that the covariance of these controllers naturally exhibits the maximum entropy characteristic, achieved without explicit incorporation of a policy entropy term in the objective as done previously. 

The contributions of this work are as follows:\\
A \textbf{control-as-inference formulation (\itwoc{})} that posits optimal control as input estimation for a dynamical system, such that the optimization objective is separated from the priors over the controls.
This allows for Bayesian inference of the controls, rather than fixing them for exploration. \\
A \textbf{practical realisation} through approximate Expectation Maximisation, performing inference via linearized Gaussian message passing in the E-Step and hyperparameter optimization in the M-Step.
Compared to previous methods, \itwoc{} has more principled regularization, relying primarily on the priors rather than heuristic methods such as line search, smoothing and annealing.
\vspace{-2mm}
\section{Input Inference for Control}
\label{sec:i2c}
Given a stochastic discrete-time fully-observed nonlinear dynamical system, $\vx_{t+1} \sim \vf(\vx_t, \vu_t)$ with state $\vx \in \mathbb{R}^{d_x}$ and input $\vu \in \mathbb{R}^{d_u}$, we wish to find the optimal control inputs $\vu^*_{0:T}$ over time horizon $T$ that minimizes the cost function $C(\vx,\vu)$ for moving from an initial state $\vx_0$ to goal state $\vx_g$.

Our proposed method reframes optimal control as inference of the inputs of the dynamical system.
This can be achieved with access to a dynamics model and by incorporating the cost function into the likelihood in an affine manner through an `observation model' $p(\vz_t|\vx_t,\vu_t)$ of our optimization variables $\vz \in \mathbb{R}^{d_z}$, such that $\alpha C(\vx,\vu){+}\beta{=}\log p(\vz_t|\vx_t,\vu_t)$.
By maximizing this likelihood
\begin{align}
	 & \max_{\vu_{0:T},\vtheta} p(\vz_{0:T}, \vx_{0:T}, \vu_{0:T}, \vtheta){=}
	p(\vx_0)
	\textstyle\prod_{t=0}^{T{\text{-}}1} p(\vx_{t+1}|\vx_{t},\vu_t)
	\textstyle\prod_{t=0}^T p(\vz_t|\vx_t,\vu_t,\vtheta)
	p(\vu_t|\vx_t),
	\label{eq:i2cmodel}
\end{align}
both the control cost (observation likelihood) and trajectory likelihood are jointly optimized, generating an estimated optimal state-action joint distribution $p(\vx, \vu)$.
From this, the conditional distribution $p(\vu|\vx)$ can be found and used as a policy.
The likelihood acts as an unconstrained control cost function by incorporating the constraint of the dynamical system, present in typical Optimal Control formulations, as an additional likelihood.
This makes sense for stochastic systems, where the dynamical system can no longer be treated as a deterministic constraint.
The likelihood also depends on hyperparameters $\vtheta$, which can be optimized via the marginal likelihood.
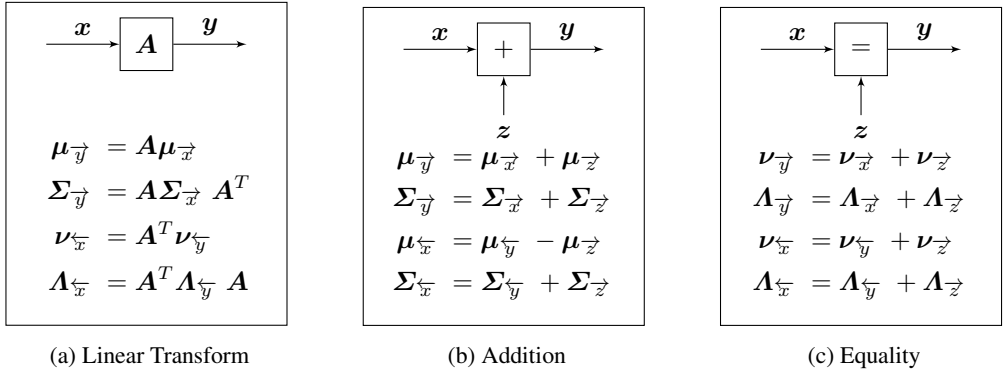
\begin{figure}[htb]
	\begin{minipage}[t][4.3cm][t]{0.32\textwidth}
		\centering
		\begin{tikzpicture}
    [%
     node distance=15mm,
     auto,>=latex',
     box/.style={draw, minimum size=0.7cm},
     short/.style={node distance=10mm},
    ]
    \tikzstyle{every node}=[font=\normalsize]
    \node [yshift=0cm, xshift=0.4cm] (start) {$~$};
    \node[box,right of=start] (A) {$\mA$} edge[<-] node[above,pos=0.5]
      {$\vx$} (start);
    \node[right of=A] (end) {$~$} edge[<-] node[above] {$\vy$}
      (A);
    \node[below of=A, yshift=-0.8cm](maths) {$
    \begin{aligned}
     \mur{y}{~}{~} &= \mA\mur{x}{~}{~} \\
     \sigr{y}{~}{~} &= \mA\sigr{x}{~}{~}\mA^T \\
     \nul{x}{~}{~} &= \mA^T\nul{y}{~}{~} \\
     \laml{x}{~}{~} &= \mA^T\laml{y}{~}{~}\mA
    \end{aligned}
    $};
    \node[box,black,inner sep=2mm,fit=(start)(A)(end)(maths)] {};
\end{tikzpicture}
	\end{minipage}
	\hfill
	\begin{minipage}[t][4.3cm][t]{0.32\textwidth}
		\centering
		\begin{tikzpicture}
    [node distance=15mm,
     auto,>=latex',
     box/.style={draw, minimum size=0.7cm},
     short/.style={node distance=10mm},
    ]
    \tikzstyle{every node}=[font=\normalsize]
    \node [yshift=0cm, xshift=0.4cm] (start) {$~$};
    \node[box,right of=start] (add) {$+$} edge[<-] node[above,pos=0.5]
      {$\vx$} (start);
    \node[short, below of=add] (z) {$~$} edge[->] node[below,pos=-0.05] {$\vz$}
      (add);
    \node[right of=add] (end) {$~$} edge[<-] node[above,pos=0.5] {$\vy$}
      (add);
    \node[below of=add, yshift=-0.8cm](maths) {$
    \begin{aligned}
     \mur{y}{~}{~} &= \mur{x}{~}{~} + \mur{z}{~}{~} \\
     \sigr{y}{~}{~} &= \sigr{x}{~}{~} + \sigr{z}{~}{~} \\
     \mul{x}{~}{~} &= \mul{y}{~}{~} -\mur{z}{~}{~}  \\
     \sigl{x}{~}{~} &= \sigl{y}{~}{~} + \sigr{z}{~}{~}
    \end{aligned}
    $};
    \node[box,black,inner sep=2mm,fit=(start)(A)(end)(maths)] {};
\end{tikzpicture}
	\end{minipage}
	\hfill
	\begin{minipage}[t][4.3cm][t]{0.32\textwidth}
		\centering
		\begin{tikzpicture}
    [node distance=15mm,
     auto,>=latex',
     box/.style={draw, minimum size=0.7cm},
     short/.style={node distance=10mm},
    ]
    \tikzstyle{every node}=[font=\normalsize]
    \node [yshift=0cm, xshift=0.4cm] (start) {$~$};
    \node[box,right of=start] (fuse) {$=$} edge[<-] node[above,pos=0.5]
      {$\vx$} (start);
    \node[short, below of=fuse] (z) {$~$} edge[->] node[below,pos=-0.05] {$\vz$}
      (A);
    \node[right of=fuse] (end) {$~$} edge[<-] node[above,pos=0.5] {$\vy$}
      (A);
    \node[below of=fuse, yshift=-0.8cm](maths) {$
    \begin{aligned}
     \nur{y}{~}{~} &= \nur{x}{~}{~} + \nur{z}{~}{~}  \\
     \lamr{y}{~}{~} &= \lamr{x}{~}{~} + \lamr{z}{~}{~} \\
     \nul{x}{~}{~} &= \nul{y}{~}{~} + \nur{z}{~}{~}  \\
     \laml{x}{~}{~} &= \laml{y}{~}{~} + \lamr{z}{~}{~}
    \end{aligned}
    $};
    \node[box,black,inner sep=2mm,fit=(start)(A)(end)(maths)] {};
\end{tikzpicture}
	\end{minipage}
	\begin{minipage}[t][.3cm][t]{0.32\textwidth}
		\vspace{-1.3mm}
		\subcaption{Linear Transform}\label{fig:linear_transform}
	\end{minipage}
	\hfill
	\begin{minipage}[t][.3cm][t]{0.32\textwidth}
		\vspace{-1.3mm}
		\subcaption{Addition}\label{fig:addition}
	\end{minipage}
	\hfill
	\begin{minipage}[t][.3cm][t]{0.32\textwidth}
		\vspace{-1.3mm}
		\subcaption{Equality}\label{fig:fusion}
	\end{minipage}
	\begin{minipage}[t][.7cm][t]{\textwidth}
		\caption{Linear Gaussian Message Passing rules for elementary state-space operations \cite{loeliger2007factor}, with the mean ($\vmu$), covariance ($\mSigma$), precision ($\mLambda=\mSigma^{-1}$) and scaled mean ($\vnu = \mLambda\vmu$), which describe the moment and information (or canonical) form of the Normal distribution respectively.}\label{fig:message_passing}
	\end{minipage}
	\vspace{-0mm}
\end{figure}
\vspace{-1mm}
\subsection{The Linear Gaussian Assumption}
\label{subsec:linear_gaussian}
By applying the linear Gaussian assumption to the models and their respective uncertainties, \Eqref{eq:i2cmodel} can not only be tackled in a tractable manner, but also compared to LQR control (Section \ref{subsec:lqr}).
Firstly, we can express the conditionals as linear state-space models,
\begin{align}
	&\text{Dynamics:} &p(\vx_{t+1}|\vx_{t},\vu_t) & : & \vx_{t+1} & = \mA_t\vx_{t} + \mB_t\vu_t + \va_t + \veta_t, & \veta_t & \sim \mathcal{N}(\mathbf{0},\;\sigEta{t}), \\
	&\text{Cost:} & p(\vz_t|\vx_t,\vu_t)     & : & \vz_t   & = \mE_t\vx_t + \mF_t\vu_t + \ve_t + \vxi_t,    & \vxi_t  & \sim \mathcal{N}(\mathbf{0},\;\sigXi).
\end{align}

Secondly, the log-likelihood is transformed into a convex function (\Eqref{eq:loglike}) which is quadratic in the optimization variables $\vx$, $\vu$ and $\vz$~\cite{Ghahramani96parameterestimation}.
\begin{align}
	{-}\mathcal{L}(\vtheta) = &
	\frac{1}{2}\textstyle\sum_{t=0}^{T-1}\log|\sigEta{t}| + 
	\frac{1}{2}\textstyle\sum_{t=0}^T(\vz_t{-} \mE_t\vx_t{-}\mF_t\vu_t{-}\ve_t)\tran
	\sigXi\inv(\vz_t{-}\mE_t\vx_t {-}\mF_t\vu_t{-}\ve_t)\notag\\
	&\hspace{-10mm}+\frac{T}{2}\log|\sigXi| +\frac{1}{2}\textstyle\sum_{t=0}^{T-1}
	(\vx_{t+1}{-}\mA_t\vx_{t}{-}\mB_t\vu_t{-}\va_t)\tran \sigEta{t}\inv(\vx_{t+1}{-}\mA_t\vx_{t}{-}\mB_t\vu_t{-}\va_t) + \dots\label{eq:loglike}
\end{align}

In \itwoc{}, the `measurement' of $\vz$ represents the desired state-action trajectory.
Therefore to transform the log-likelihood of $\vz$ to a quadratic control cost,
the precision of the `observation noise' $\vxi$ is $\sigXi^{-1}{=}\lamXi{=}\alpha \bm{\Theta}$, where $\bm{\Theta}$ represents the weights of the cost function and $\alpha$ accounts for its scale invariance.
For the standard LQ problem (Section \ref{subsec:lqr}), $\vz_t{=}\left[\vx_g \; \vu_g\right]^\intercal$ and $\bm{\Theta}{=}\text{diag}(\mQ, \mR)$.
Our hyperparameters $\vtheta$ include $\alpha$, the scale factor, along with the priors over the inputs $\vu$.
In \Eqref{eq:loglike}, $\alpha$ acts as the scale factor of the LQ cost against the other terms in the likelihood.
Typically for multi-objective cost functions this scaling must be user-defined, but as it has a probabilistic interpretation here, it can be iteratively estimated during inference.   
As $\alpha$ scales the given control cost $\bm{\Theta}$ such that it can be used as the observation noise precision $\lamXi$, it can be estimated based on the current estimated state-action trajectory deviation about the goal.
This inference is carried out using the Expectation Maximization (EM) algorithm \cite{dempster1977maximum}, treating $\alpha$ as a latent variable.
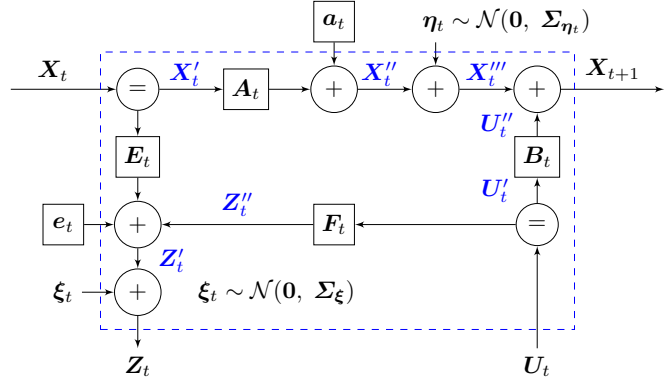
\begin{wrapfigure}[17]{r}{0.63\textwidth}
	\vspace{-1mm}
	\begin{tikzpicture}
[%
 node distance=20mm,
 auto,>=latex',
 box/.style={draw, minimum size=0.6cm},
 short/.style={node distance=10mm},
 transform canvas={scale=0.9} %
]
\node [yshift=-1.1cm] (start) {}; %
\node[circle, draw, right of=start] (x_fuse) {$=$} edge[<-] node[above,pos=0.6] {{$\mX_{t}$}} (start);
\node[box, below of=x_fuse, yshift=1cm] (E) {$\mE_t$} edge[<-](x_fuse);
\node[circle, draw, below of=E, yshift=1cm] (z_add)  {$+$} edge[<-] node {} (E);
\node[box, left of=z_add, xshift=0.9cm] (E_off) {$\ve_t$} edge[->](z_add);
\node[circle, draw, below of=z_add, yshift=1cm] (xi_add)  {$+$} edge[<-] node[right=5pt,pos=0.4] {\textcolor{blue}{$\mZ_t'$}}(z_add);
\node[left of=xi_add, xshift=0.9cm] (xi) {$\bm{\xi}_t$} edge[->](xi_add);
\node[short, right of=xi_add] (sig_xi) {\hspace{21mm}$\bm{\xi}_t \sim \mathcal{N}(\mathbf{0},\;\sigXi)$};

\node[short, below of=xi_add, yshift=-0.1cm] (z) {\hspace{0mm}$\mZ_t$} edge[<-] node {} (xi_add);
\node[box,right of=x_fuse, xshift=-0.4cm] (A) {$\mA_t$} edge[<-] node[above=-2pt,pos=0.6] {\textcolor{blue}{$\mX_{t}'$}} (x_fuse);
\node[circle, draw, right of=A,xshift=-0.7cm] (A_add) {$+$} edge[<-] (A);
\node[box, short, above of=A_add] (A_off) {$\va_t$} edge[->] (A_add);
\node[circle, draw, right of=A_add,xshift=-0.5cm] (sigv_add) {$+$} edge[<-] node[above=-2pt,pos=0.6] {\textcolor{blue}{$\mX_t''$}} (A_add);
\node[short, above of=sigv_add] (sig_v) {\hspace{21mm}$\veta_t \sim \mathcal{N}(\mathbf{0},\;\sigEta{t})$} edge[->] (sigv_add);
\node[circle, draw, right of=sigv_add, xshift=-0.5cm] (u_add) {$+$} edge[<-] node[above=-2pt] {\textcolor{blue}{$\mX_t'''$}} (sigv_add);
\node[right of=u_add] (end) {} edge[<-] node[above] {$\mX_{t+1}$}
  (u_add); %
\node[box, below of=u_add, yshift=1cm] (B) {$\mB_t$} edge[->] node[left=6pt] {\textcolor{blue}{$\mU_t''$}} (u_add);
\node[circle, draw, below of=B, yshift=1cm] (u_fuse) {$=$} edge[->] node[left=8pt] {\textcolor{blue}{
		$\mU_t'$}} (B);
\node[box, left of=u_fuse, xshift=-1.0cm] (F) {$\mF_t$} edge[<-](u_fuse); 
\draw[] (F)edge[->] node[above] {\textcolor{blue}{$\mZ_t''$}} (z_add);
\node[short, below of=u_fuse, yshift=-1.1cm] (u) {\hspace{-0mm}$\mU_t$} edge[->] node[pos=0.2] {} (u_fuse);
\node[box,blue,dashed,inner sep=2mm,fit=(A)(A_add)(u_add)(sigv_add)(x_fuse)(B)(u_fuse)(F)(z_add)(xi_add)] {};
\end{tikzpicture}
	\vspace{47mm}
	\caption{Forney factor graph of the linear Gaussian dynamical system used by \itwoc{}. Blue terms are intermediate variables used in the message derivations (Section \ref{subsec:msg}).}
	\label{fig:FactorGraph}
\end{wrapfigure}

\textbf{Expectation Step} \\
The E-Step, estimating the state-action trajectory, can be performed in a tractable manner through linear Gaussian message passing. For model-based signal processing on linear Gaussian state space models ~\cite{loeliger2007factor,bruderer2015input,loeliger2016sparsity}, expressing inference problems as Forney-style factor graphs enables the construction of message-passing algorithms by following straightforward rules (see Figure \ref{fig:message_passing}).
For cycle-free graphs, the messages can be expressed in closed-form.
The forward messages (i.e. $\overrightarrow{\vx}$) represents the priors, while the backward messages (i.e. $\overleftarrow{\vx}$) represent likelihood functions (up to a scale factor).
The updated belief is the posterior of an edge, which are the product of the edge's forward and backward message:
\begin{align}
	\sigm{x}{~}{~} & = (\!\lamr{x}{~}{~}\!+\!\laml{x}{~}{~}\!)^{-1}, \;\;\; \mum{x}{~}{~}\!=\! \sigm{x}{~}{~}(\nur{x}{~}{~}\!+\!\nul{x}{~}{~}) \label{eq:marginal}.
\end{align}

In \itwoc{}, the backward messages perform optimal control, so the posterior states and controls represent a regularized update of the estimated optimal state-action trajectory.

\begin{wrapfigure}[25]{r}{0.5\textwidth}
	\begin{minipage}{0.5\textwidth}
		\begin{algorithm}[H]
			\KwData{$T$, $\alpha$, $\delta_\alpha$, $\vf(\vx, \vu)$, $\vg(\vx, \vu)$ \\
				$\mur{x}{~}{0}, \sigr{x}{~}{0}, \mur{u}{~}{t}, \sigr{u}{~}{t}$ for $t=0\!:\!T$}
			\KwResult{$\mK_t,\vk_t,\mSigma_{k_t}$ for $t=0:T$}
			\While{not converged}{
				\texttt{// E-Step}\\
				
				\For{$i\gets0$ \KwTo $T-1$}
				{
					Compute $\mur{x}{~}{t+1}$, $\sigr{x}{~}{t+1}$ from \\
					forward messages (\Eqref{eq:fwd_start}-\ref{eq:fwd_end}),\\
					updating $\mA_t$, $\va_t$,$\mB_t$,$\mE_t$, $\ve_t$ and $\mF_t$
				}
				\For{$i\gets T$ \KwTo $1$}
				{
					Compute $\mum{x}{~}{t}$, $\sigm{x}{~}{t}$
					$\mum{u}{~}{t}$, $\sigm{u}{~}{t}$ from \\
					backward messages and \\
					marginalisation (\Eqref{eq:bwd_start}-\ref{eq:bwk_end})
				}
				\texttt{// M-Step}\\
				Update $\alpha$ with reg. (\Eqref{eq:alpha_end}, \ref{eq:alpha_reg}) \\
				Update priors, $\mur{u}{~}{}{=}\mum{u}{~}{}$,
				$\sigr{u}{~}{}{=}\sigm{u}{~}{}$
			}
			\texttt{// Controller}\\
			Computer linear Gaussian controller\\ $\mK_t,\vk_t,\mSigma_{k_t}$ for $t=0\!:\!T$
			from messages\\ (\Eqref{eq:K_t}-\ref{eq:sigma_k})\\
			\caption{EM for Linear Gaussian \itwoc{}}
			\vspace{2mm}
		\end{algorithm}
	\end{minipage}
\end{wrapfigure}
The message-passing on the graph of Figure~\ref{fig:FactorGraph} performs the same inference as Kalman filtering and smoothing \cite{anderson2012optimal}, with the addition that the inputs are also uncertain\footnote{If the input is incorporated into the state, the two procedures become identical, however the joint dynamics then become degenerate due to the independence of the inputs}
Additionally, the inference starts with an `innovation' (observation) of $\vx_0$ in order to evaluate $(\vx_t,\vu_t)$ rather than $(\vx_{t+1},\vu_t)$, but this is a minor discrepancy as the subsequent prediction and innovation steps are the same.
The forward and backward messages are derived in Sections \ref{subsec:forward_msg}-\ref{subsec:backward_msg}.
While the message-passing form is more verbose than the standard Kalman filtering and smoothing equations, they allow us to appreciate how this framework performs optimal control \cite{hoffmann2017linear}.
From \Eqref{eq:loglike} with the LQ-equivalent $\vz_t$ and $\bm{\Theta}$, it is clear that the negative log-likelihood acts an upper bound on the LQ cost, as it incorporates the trajectory likelihood, which depends on the system's stochasticity and uncertainty in controls.
Therefore, as the EM algorithm maximizes the log-likelihood, it in turn minimizes the LQ cost, performing Bayesian optimal control.
The further connections between \itwoc{} and LQ control are discusses in Section \ref{subsec:connections} and \ref{subsec:lqr_equiv}.

\textbf{Maximisation Step}\\
To update $\alpha$, the scale factor between the LQ cost $\bm{\Theta}$ and the estimated $\lamXi$ must be found. 
This is derived by maximizing the expected log-likelihood via the derivative:
\begin{align}
	-2\frac{\partial}{\partial \alpha}\mathbb{E}[\mathcal{L}(\alpha)] & = \frac{\partial}{\partial \alpha}(\tr\{\sigXi^{-1}\hat{\sigXi}\} +  T\log|\sigXi|)
	= -\tr\{\bm{\Theta}\hat{\sigXi}\} +  T d_z\alpha^{-1} = 0, \label{eq:alpha_end}                                                                                                                                                                                                                                               \\
	\text{where} \qquad \hat{\sigXi}                                  & = \textstyle\sum_{t=0}^T\left[(\vz_t - \mE_t\mum{x}{~}{t}\!-\!\mF_t\mum{u}{~}{t})(\vz_t\!-\!\mE_t\mum{x}{~}{t}\!-\!\mF_t \mum{u}{~}{t})^\intercal\!+\!\mE_t\sigm{x}{~}{t}\mE_t^\intercal\!+\!\mF_t\sigm{u}{~}{t}\mF_t^\intercal\right]. \nonumber
\end{align}
In practice this means that over EM iterations, as the state-action trajectory moves towards the goal, $\lamXi$ and therefore $\alpha$ steadily increases.
This in turn results in the control cost term increasing in significance in the log-likelihood (\Eqref{eq:loglike}).
The resulting annealing effect aids in stabilizing the optimization.
This effect bares a resemblance to curriculum learning \cite{bengio2009curriculum}, where the task (e.g. cost function) increases in difficulty as the performance improves, as a strategy for learning complex tasks effectively.

\textbf{Linear Gaussian Controller} \\
For finite horizon LQ control, it can be shown that a time-varying linear controller is the optimal policy.
Here we show that this is true for the inference setting as well.
By examining the conditional distribution between the marginalized posteriors of $\vx$ and $\vu$ at each timestep, a time-varying linear Gaussian controller (\Eqref{eq:K_t}-\ref{eq:sigma_k}) can be derived from the messages (see Section \ref{subsec:controller_msg}).
For a time-varying linear Gaussian controller of the form $\vu_t \sim \mathcal{N}(\mK_t \vx_t + \mK_t,\;\sigm{k}{~}{t})$, \itwoc{} computes the parameters as
\begin{align}
	\mK_t          & = -\sigm{u}{~}{t}\mB_t\mGamma_{t+1} \laml{x}{~}{t+1}\mPsi_{t+1}\mA_t,\label{eq:K_t}                                                                                                        \\
	\vk_t          & = \sigm{u}{~}{t}(\nur{u}{~}{t} + \mF_t^\intercal(\sigXi + \mE_t\sigr{x}{~}{t}\mE_t\tran)\inv(\vz_t-\mE_t\mur{x}{}{t}-\ve_t)                                                                \notag\\
	               & \hspace{6mm}+\mB_t^\intercal(
	\mGamma_{t+1}\nul{x}{~}{t+1}
	+(\mI - \mGamma_{t+1})\nur{x}{''}{t}
	-\mGamma_{t+1}\laml{x}{~}{t+1}\mPsi_{t+1}\va_t)),\label{eq:k_t}                                                                                                                                             \\
	\sigm{k}{~}{t} & = \sigm{u}{~}{t} =  (\lamr{u}{~}{t} + \mF_t^\intercal(\sigXi + \mE_t\protect\sigr{x}{~}{t}\mE_t\tran)\inv\mF_t + \mB_t^\intercal\mGamma_{t+1}\laml{x}{~}{t+1}\mB_t)^{-1}.\label{eq:sigma_k}
\end{align}
In Section \ref{subsec:connections}, it is shown how the expressions for the controller resemble the corresponding expressions for LQ control.
Moreover, the discrepancy between the \itwoc{} and LQ controllers can be interpreted as uncertainty-derived regulation.
In the \itwoc{} controller two additional (dimensionless) terms appear, $\mGamma$ and $\mPsi$ (see Section \ref{subsec:controller_msg}), which are functions of $\sigl{x}{}{t+1}$, $\sigr{x}{''}{t}$ and $\sigr{u}{''}{t}$.
As process uncertainty increases, $\mGamma$ acts to `turn off' the optimal control terms of the controller and rely on the priors.
Meanwhile, $\mPsi$ represents the confidence in the controller, which counteracts the attenuating effects of $\mGamma$ given sufficient control certainty.
These findings parallel the `turn-off phenomenon' observed in Dual Control \cite{aoki1967optimization, bar1981stochastic} and Bayesian Reinforcement Learning \cite{DBLP:journals/jmlr/KlenskeH16}, where actions are attenuated under uncertainty.
This behaviour is important for settings such as probabilistic Model-based Reinforcement Learning \cite{deisenroth2013survey}, where localised regions of uncertainty can indicate modelling error, and such errors can lead to detrimental policy updates.
Attenuating the policy updates in these regions between model learning iterations would mitigate this pitfall.
\subsubsection{Connections to Finite Horizon Maximum Entropy LQR}
\label{subsec:connections}
To understand how this framework performs optimal control, we look at the backward messages of the probabilistic graphical model described in Figure \ref{fig:FactorGraph} with a control perspective \cite{hoffmann2017linear}.
By looking at the backward messages of the state $\mathbold{{\overleftarrow{X}_{t}}}$ in Section \ref{subsec:ric_msg}, the backwards evolution of the precision (\Eqref{eq:lam_ric}) and scaled-mean (\Eqref{eq:nu_ric}) can be seen to have a similar Ricatti form to the quadratic value function parameters for LQ control (\Eqref{eq:P_lqr}-\ref{eq:p_lqr}).
Extending this analysis to find the linear Gaussian controllers from the conditional distributions, we see that some of the equivalent terms have the additional uncertainty-weighted scalar term $\mGamma$. Table \ref{tab:equiv} details the correspondence.
	{\renewcommand{\arraystretch}{1}
		\begin{table}[t]
			\centering
			\begin{tabular}{>{\centering\arraybackslash}m{1cm} >{\centering\arraybackslash}m{4cm} >{\centering\arraybackslash}m{5cm} @{}m{0pt}@{}}
				\toprule
				LQR                                                                        & {Riccati Backward Message}                                                 & {Message-derived Controller} \\
				\midrule
				$\mQ$                                                                      &
				$\mE_t^\intercal(\sigXi + \mF_t\protect\sigr{u}{~}{t}\mF_t\tran)\inv\mE_t$ & $\mE_t^\intercal(\sigXi + \mF_t\protect\sigr{u}{~}{t}\mF_t\tran)\inv\mE_t$                                \\
				$\mR$                                                                      &
				$\mF_t^\intercal(\sigXi + \mE_t\protect\sigr{x}{~}{t}\mE_t\tran)\inv\mF_t$ &
				$\mF_t^\intercal(\sigXi + \mE_t\protect\sigr{x}{~}{t}\mF_t\tran)\inv\mF_t$                                                                                                             \\
				$\mP_t$                                                                    & $\laml{x}{~}{t}$                                                           & $\mGamma_t\laml{x}{~}{t}$    \\
				$\vp_t$                                                                    & $-\nul{x}{~}{t}$                                                           & $-\mGamma_t\nul{x}{~}{t}$    \\
				\bottomrule\vspace{2mm}
			\end{tabular}
			\caption{Due to the formulation of \itwoc{}, the precision of the observation noise is proportional to the LQ cost function weights. Additionally, due to the linear Gaussian assumption, we can show that the precision and scaled-mean of the backward messages of the state belief correspond to the value function parameters in LQR. These equivalences are explained further in Section \ref{subsec:lqr_equiv}.}
			\label{tab:equiv}
			\vspace{-0.5cm}
		\end{table}
	}

The control covariance, \Eqref{eq:sigma_k}, can be seen to resemble that of a Maximum Entropy controller.
In Control Theory and Reinforcement Learning, the entropy of a policy can be interpreted as a metric for robustness, so a maximum entropy objective has been added to cost functions as regularisation~\cite{ziebart2010modeling}.
Augmenting the LQ cost function with the entropy of the control inputs, the covariance of the input at each timestep can be shown to be $\mSigma_t = (\mR + \mB^\intercal \mP_{t+1} \mB)^{-1}$ (using LQ notation, see Section \ref{subsec:lqr}) \cite{levine2013variational}. Comparing this to \Eqref{eq:sigma_k} and Table \ref{tab:equiv}, it can be seen that this maximum entropy control is calculated by the backward message, and combined with the prior (forward message) to construct the posterior. This fusion is important as the prior can be used to regularize exploration during inference, which is essential for mitigating the effects of linearizing the dynamics during approximate inference of nonlinear systems (see Section \ref{subsec:nonlinear}). This smoothing mechanism has previously been added explicitly or via constraints on the trajectory update during optimization.
\subsection{Nonlinear \itwoc{} through Approximate Inference}
\label{subsec:nonlinear}
The linear analysis conducted here can be naturally extended to nonlinear dynamical systems through linearization, taking the Jacobian of the dynamics and observation models about the current state-action trajectory.
This approach has been applied to both state estimation (i.e. Extended Kalman Smoothing) and optimal control (i.e. DDP).
From a probabilistic perspective, this linearization renders the inference approximate.
As a consequence, careful consideration of the priors and additional regularization is required, as the act of linearizing imposes a requirement of \textit{local} improvement during inference.
Placing small priors on $\vu$ ensures that the Bayesian posterior remains close to the prior, and was found to be critical for systems that where highly nonlinear or with low sampling frequencies.
As in Extended Kalman Filtering and other inference schemes for nonlinear systems \cite{ghahramani1999learning}, the dynamics are linearized in the forward pass.
This linearization-based approximate inference can be viewed as Gauss-Newton optimization \cite{bell1994iterated}, making it closely related to approximate trajectory optimization algorithms such as iLQR.
Additionally, it was found that the $\alpha$ update during the M-Step must be restricted to ensure the state-action distribution did not change significantly between iterations.
By looking at a bound $\delta_\xi$ on the KL divergence between $\mZ$ updates (\Eqref{eq:alpha_reg}), this in fact can be applied as a bound $\delta_\alpha$ on the update ratio. As the expression 
\begin{align}
	\kld{\!\mZ^{i}}{\mZ^{i+1}}\!=
	\!\frac{1}{2}\!\left[\log\frac{|\sigXi^{i+1}|}{|\sigXi^{i}|} \!+\!\tr\{\lamXi^{i+1}\sigXi^{i}\}\!-\!d_z\right]\!=\!   \frac{1}{2}\!\left[\log\frac{\alpha^{i+1}}{\alpha^{i}}\!+\!d_z\frac{\alpha^{i}}{\alpha^{i+1}}\!-\!d_z\right]\! & \leq\!\delta_\xi\label{eq:alpha_reg}
\end{align}
is monotonic increasing in the ratio $\nicefrac{\alpha^{i+1}}{\alpha^{i}}$. %
From the perspective of approximate EM, the regularized M-Step is motivated by mitigating the adverse effect of the linearization assumption on the likelihood estimate. \cite{yi2015regularized}.
\section{Experimental Results}
\label{sec:result}
An empirical evaluation is presented, first to highlight the equivalence of {\itwoc} to the LQR solution and second to compare {\itwoc} to state-of-the-art algorithms on nonlinear dynamical systems\footnote{The code is available at \url{https://github.com/JoeMWatson/input-inference-for-control}}.
\subsection{Equivalence with finite-horizon LQR by Dynamic Programming}
\label{subsec:lqr_experiment}
In Section \ref{subsec:linear_gaussian}, the LQR problem was used to motivate the linear Gaussian assumption for \itwoc{}.
In Section \ref{subsec:lqr_equiv} it is shown how, under specific settings, the message passing expressions reduce to those found when solving the LQR problem via Dynamic Programming. Figure \ref{fig:lqr_equiv} illustrates this numerically, for an LQR problem described in Section {\ref{subsec:lqr_experiment_details}.
\begin{figure}[t!]
	\begin{minipage}[t][5.8cm][t]{0.49\textwidth}
		\centering
		\input{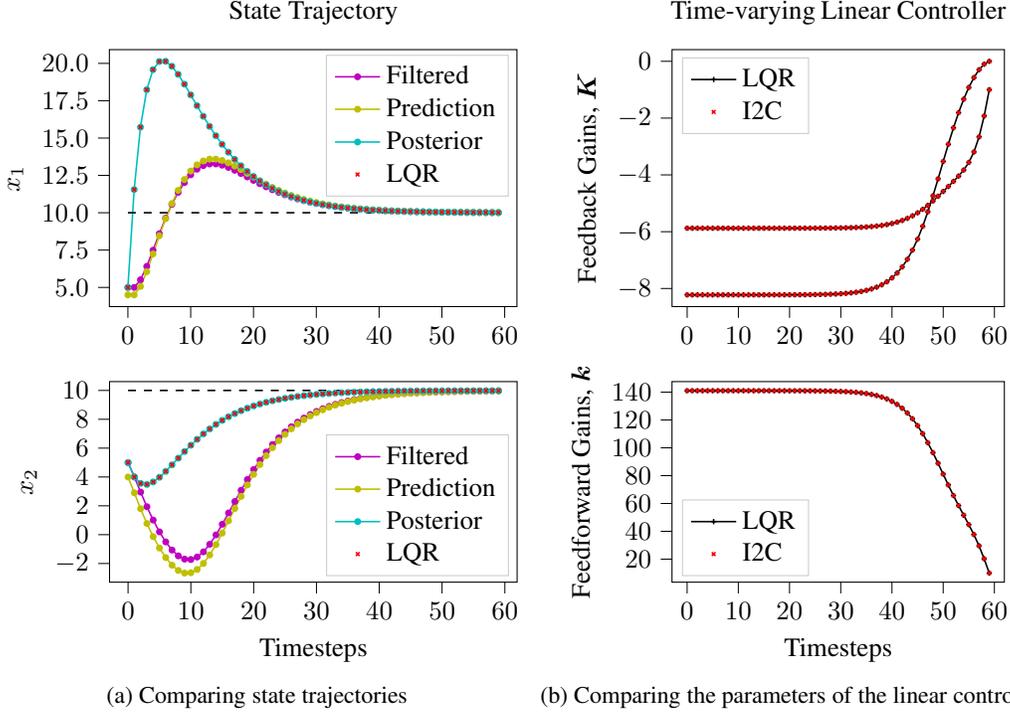}
	\end{minipage}
	\begin{minipage}[t][5.8cm][t]{0.49\textwidth}
		\centering
		\begin{tikzpicture}

\begin{groupplot}[group style={group size=1 by 2}]
\nextgroupplot[
width=6cm,
height=5cm,
legend cell align={left},
legend style={at={(0.03,0.97)}, anchor=north west, draw=white!80.0!black},
tick align=outside,
tick pos=left,
title={Time-varying Linear Controller},
x grid style={white!69.01960784313725!black},
xmin=-2.95, xmax=61.95,
xtick style={color=black},
xtick={-10,0,10,20,30,40,50,60,70},
xticklabels={\(\displaystyle -10\),\(\displaystyle 0\),\(\displaystyle 10\),\(\displaystyle 20\),\(\displaystyle 30\),\(\displaystyle 40\),\(\displaystyle 50\),\(\displaystyle 60\),\(\displaystyle 70\)},
y grid style={white!69.01960784313725!black},
ylabel={Feedback Gains, \(\displaystyle \mathbold{{K}}\)},
ymin=-8.63683105034066, ymax=0.411277669063841,
ytick style={color=black},
ytick={-10,-8,-6,-4,-2,0,2},
yticklabels={\(\displaystyle -10\),\(\displaystyle -8\),\(\displaystyle -6\),\(\displaystyle -4\),\(\displaystyle -2\),\(\displaystyle 0\),\(\displaystyle 2\)}
]
\addplot [semithick, black, mark=+, mark size=1, mark options={solid}]
table {%
0 -5.87778697521724
1 -5.87778610229459
2 -5.87778494976457
3 -5.87778342807112
4 -5.87778141897713
5 -5.87777876638086
6 -5.87777526419276
7 -5.87777064033233
8 -5.87776453560426
9 -5.87775647581662
10 -5.87774583498145
11 -5.87773178674916
12 -5.87771324032071
13 -5.8776887558855
14 -5.87765643305948
15 -5.87761376372687
16 -5.87755743796759
17 -5.87748308817804
18 -5.87738495180705
19 -5.87725542699336
20 -5.87708448737808
21 -5.87685891192768
22 -5.87656127205786
23 -5.87616860085364
24 -5.87565064672834
25 -5.87496758529118
26 -5.87406702724961
27 -5.87288011567641
28 -5.87131645214008
29 -5.86925752827459
30 -5.86654826975498
31 -5.86298622987219
32 -5.85830791386471
33 -5.8521716993455
34 -5.84413688950855
35 -5.83363867284724
36 -5.81995928986168
37 -5.80219670631228
38 -5.77923380760699
39 -5.74971383271115
40 -5.71203165025674
41 -5.66435539718801
42 -5.60469798793055
43 -5.53106059227638
44 -5.44166578938174
45 -5.33528032913137
46 -5.21159064525642
47 -5.07153948521319
48 -4.91747391657354
49 -4.75292220709799
50 -4.58183602623048
51 -4.40719826938688
52 -4.22894841606382
53 -4.04114951035288
54 -3.82826337277279
55 -3.56071492480243
56 -3.19170251681349
57 -2.66202739319328
58 -1.92568125516102
59 -1
};
\addlegendentry{LQR}
\addplot [semithick, red, mark=x, mark size=1, mark options={solid}, only marks]
table {%
0 -5.87778686671388
1 -5.87780555379253
2 -5.87778484126114
3 -5.87778331956765
4 -5.87778131047359
5 -5.87777865787726
6 -5.87777515568904
7 -5.87777053182845
8 -5.87776442710018
9 -5.87775636731225
10 -5.87774572647668
11 -5.87773167824387
12 -5.87771313181469
13 -5.87768864737851
14 -5.87765632455118
15 -5.8776136552168
16 -5.87755732945512
17 -5.87748297966233
18 -5.87738484328699
19 -5.87725531846743
20 -5.87708437884425
21 -5.87685880338326
22 -5.87656116349921
23 -5.87616849227594
24 -5.87565053812514
25 -5.87496747665391
26 -5.87406691856688
27 -5.87288000693316
28 -5.87131634331641
29 -5.86925741934432
30 -5.86654816068374
31 -5.86298612061525
32 -5.85830780436396
33 -5.85217158952617
34 -5.84413677927516
35 -5.83363856207932
36 -5.81995917840914
37 -5.80219659399176
38 -5.77923369419956
39 -5.74971371796372
40 -5.71203153388956
41 -5.66435527891136
42 -5.60469786747419
43 -5.53106046943372
44 -5.44166566406705
45 -5.33528020144105
46 -5.21159051551505
47 -5.07153935397456
48 -4.9174737845421
49 -4.75292207494529
50 -4.58183589430463
51 -4.40719813734742
52 -4.2289482825124
53 -4.04114937258847
54 -3.82826322694458
55 -3.56071476707126
56 -3.19170234663445
57 -2.66202721963538
58 -1.92568110363848
59 -0.999999909090918
};
\addlegendentry{I2C}
\addplot [semithick, black, mark=+, mark size=1, mark options={solid}, forget plot]
table {%
0 -8.22536251310138
1 -8.22535938749397
2 -8.22535526070461
3 -8.22534981205528
4 -8.22534261816431
5 -8.22533312005868
6 -8.22532057975828
7 -8.22530402296318
8 -8.22528216339895
9 -8.22525330295417
10 -8.22521519987133
11 -8.22516489478297
12 -8.22509848112968
13 -8.22501080220792
14 -8.22489505144876
15 -8.22474224509557
16 -8.22454052667579
17 -8.22427424982103
18 -8.22392276914302
19 -8.22345884680262
20 -8.222846553557
21 -8.22203850545212
22 -8.22097222844948
23 -8.21956538005523
24 -8.2177094757393
25 -8.21526166427049
26 -8.21203396533499
27 -8.20777922036268
28 -8.20217281003636
29 -8.19478895962796
30 -8.18507019364668
31 -8.17228823622155
32 -8.15549443110775
33 -8.13345766787256
34 -8.10458801478107
35 -8.06684505467191
36 -8.01763174488468
37 -7.95367814248757
38 -7.87092545572093
39 -7.76443064727187
40 -7.62832604502112
41 -7.45588682376748
42 -7.23977875043196
43 -6.97257098831894
44 -6.64758841817649
45 -6.26012292416003
46 -5.80890458784688
47 -5.29755768715906
48 -4.73559064996626
49 -4.13840876096099
50 -3.52601292005029
51 -2.92047268363891
52 -2.34277359305527
53 -1.80996976645877
54 -1.33356287767856
55 -0.919766866346656
56 -0.571985171734712
57 -0.295193465942433
58 -0.0999174236168456
59 0
};
\addplot [semithick, red, mark=x, mark size=1, mark options={solid}, only marks, forget plot]
table {%
0 -8.22536226981108
1 -8.22555338127682
2 -8.22535501741431
3 -8.22534956876503
4 -8.22534237487403
5 -8.22533287676845
6 -8.22532033646803
7 -8.22530377967289
8 -8.22528192010863
9 -8.22525305966369
10 -8.22521495658065
11 -8.22516465149201
12 -8.22509823783821
13 -8.22501055891573
14 -8.22489480815552
15 -8.2247420018008
16 -8.22454028337878
17 -8.22427400652078
18 -8.22392252583822
19 -8.22345860349136
20 -8.22284631023666
21 -8.22203826211914
22 -8.22097198509889
23 -8.21956513668035
24 -8.2177092323309
25 -8.2152614208162
26 -8.212033721818
27 -8.20777897676057
28 -8.20217256631915
29 -8.19478871575592
30 -8.18506994956738
31 -8.17228799186679
32 -8.15549418638925
33 -8.13345742267785
34 -8.10458776896901
35 -8.06684480806932
36 -8.01763149728466
37 -7.95367789365348
38 -7.87092520539827
39 -7.7644303952148
40 -7.62832579104135
41 -7.45588656781413
42 -7.23977849271169
43 -6.97257072945939
44 -6.64758815942357
45 -6.26012266757724
46 -5.80890433645498
47 -5.29755744493007
48 -4.73559042158794
49 -4.13840855133719
50 -3.52601273359969
51 -2.92047252356072
52 -2.34277346078585
53 -1.80996966150931
54 -1.33356279793646
55 -0.919766808785103
56 -0.571985133291813
57 -0.295193444011187
58 -0.0999174152835178
59 -2.98645027035372e-17
};

\nextgroupplot[
width=6cm,
height=4.25cm,
legend cell align={left},
legend style={draw=white!80.0!black, at={(0.03,0.03)}, anchor=south west},
tick align=outside,
tick pos=left,
x grid style={white!69.01960784313725!black},
xlabel={Timesteps},
xmin=-2.95, xmax=61.95,
xtick style={color=black},
xtick={-10,0,10,20,30,40,50,60,70},
xticklabels={\(\displaystyle -10\),\(\displaystyle 0\),\(\displaystyle 10\),\(\displaystyle 20\),\(\displaystyle 30\),\(\displaystyle 40\),\(\displaystyle 50\),\(\displaystyle 60\),\(\displaystyle 70\)},
y grid style={white!69.01960784313725!black},
ylabel={Feedforward Gains, \(\displaystyle \mathbold{{k}}\)},
ymin=3.44840556075125, ymax=147.583463224225,
ytick style={color=black},
ytick={0,20,40,60,80,100,120,140,160},
yticklabels={\(\displaystyle 0\),\(\displaystyle 20\),\(\displaystyle 40\),\(\displaystyle 60\),\(\displaystyle 80\),\(\displaystyle 100\),\(\displaystyle 120\),\(\displaystyle 140\),\(\displaystyle 160\)}
]
\addplot [semithick, black, mark=+, mark size=1, mark options={solid}]
table {%
0 141.031494883186
1 141.031454897885
2 141.031402104692
3 141.031332401264
4 141.031240371414
5 141.031118864395
6 141.03095843951
7 141.030746632955
8 141.030466990032
9 141.030097787708
10 141.029610348528
11 141.028966815321
12 141.028117214504
13 141.026995580934
14 141.025514845082
15 141.023560088224
16 141.020979646434
17 141.017573379991
18 141.013077209501
19 141.00714273796
20 140.999310409351
21 140.988974173798
22 140.975335005073
23 140.957339809089
24 140.933601224676
25 140.902292495617
26 140.861009925846
27 140.806593360391
28 140.734892621764
29 140.640464879025
30 140.516184634016
31 140.352744660937
32 140.138023449724
33 139.85629367218
34 139.487249042896
35 139.004837275191
36 138.375910347463
37 137.558748487998
38 136.501592633279
39 135.14144479983
40 133.403576952779
41 131.202422209555
42 128.444767383625
43 125.036315805953
44 120.892542075582
45 115.954032532914
46 110.204952331033
47 103.690971723723
48 96.530645665398
49 88.9133096805898
50 81.0784894628077
51 73.2767095302578
52 65.7172200911909
53 58.5111927681166
54 51.6182625045135
55 44.8048179114909
56 37.6368768854821
57 29.5722085913571
58 20.2559867877787
59 10
};
\addlegendentry{LQR}
\addplot [semithick, red, mark=x, mark size=1, mark options={solid}, only marks]
table {%
0 141.03149136525
1 141.031869694068
2 141.031398586755
3 141.031328883327
4 141.031236853476
5 141.031115346457
6 141.03095492157
7 141.030743115013
8 141.030463472088
9 141.030094269759
10 141.029606830573
11 141.028963297358
12 141.028113696529
13 141.026992062942
14 141.025511327067
15 141.023556570176
16 141.020976128339
17 141.017569861831
18 141.013073691252
19 141.007139219588
20 140.999306890809
21 140.988970655024
22 140.975331485981
23 140.957336289563
24 140.933597704561
25 140.902288974701
26 140.861006403849
27 140.806589836937
28 140.734889096356
29 140.640461351002
30 140.516181102511
31 140.35274112482
32 140.138019907532
33 139.85629012204
34 139.487245482442
35 139.004833701486
36 138.375906756938
37 137.558744876453
38 136.501588995978
39 135.141441131785
40 133.403573249309
41 131.202418467255
42 128.444763601859
43 125.036311988931
44 120.892538234906
45 115.954028690183
46 110.2049485197
47 103.690967989046
48 96.5306420613003
49 88.9133062628248
50 81.0784862790432
51 73.2767066090814
52 65.7172174329825
53 58.5111903409778
54 51.6182602488104
55 44.8048157585636
56 37.6368747992626
57 29.5722066364656
58 20.25598518922
59 9.99999909090917
};
\addlegendentry{I2C}
\end{groupplot}

\end{tikzpicture}
	\end{minipage}
	\begin{minipage}[t][0.3cm][t]{0.49\textwidth}
		\vspace{-2mm}
		\subcaption{Comparing state trajectories}.\label{fig:lqr_traj}
		\vfill
	\end{minipage}
	\hfill
	\begin{minipage}[t][0.3cm][t]{0.49\textwidth}
		\vspace{-2mm}
		\subcaption{Comparing the parameters of the linear controller}\label{fig:lqr_controller}
	\end{minipage}
	\begin{minipage}[t][1cm][t]{\textwidth}
		\caption{Demonstrating how \itwoc{} generalizes the Dynamic Programming Finite Horizon LQR solution. This is achieved when the controls have a large prior and the certainty in the target observation is high.
		Note `Filtered' and `Prediction' correspond to $\protect\mur{x}{'}{t}$ and $\protect\mur{x}{~}{t+1}$ in Figure \ref{fig:FactorGraph} respectively. }\label{fig:lqr_equiv}
	\end{minipage}
	\vspace{-3mm}
\end{figure}

\begin{figure}[ht]
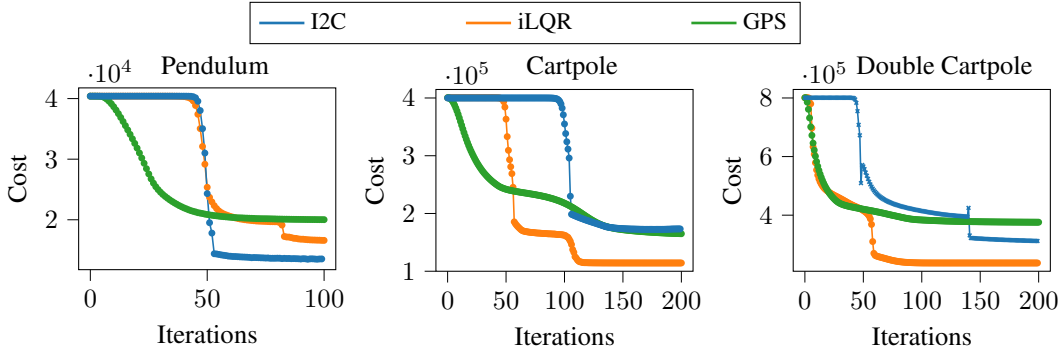

	\begin{minipage}[t]{\columnwidth}
		\centering
		\begin{tikzpicture}

\definecolor{color0}{rgb}{0.12156862745098,0.466666666666667,0.705882352941177}
\definecolor{color1}{rgb}{1,0.498039215686275,0.0549019607843137}
\definecolor{color2}{rgb}{0.172549019607843,0.627450980392157,0.172549019607843}

\begin{axis}[
hide axis,
width=14cm,
xmin=10, xmax=50,
ymin=0, ymax=1.0,
legend cell align={center},
legend columns=3,
legend style={/tikz/every even column/.append style={column sep=1.5cm}}]
]

\addlegendimage{no markers, line width=1.0pt, color0}
\addlegendentry{\small I2C};
\addlegendimage{no markers, line width=1.0pt, color1}
\addlegendentry{\small iLQR };
\addlegendimage{no markers, line width=1.0pt, color2}
\addlegendentry{\small GPS};

\end{axis}

\end{tikzpicture}
	\end{minipage}
	\hfill
	\begin{minipage}[b][3.7cm][t]{0.32\textwidth}
		\centering
		\begin{tikzpicture}

\definecolor{color0}{rgb}{0.12156862745098,0.466666666666667,0.705882352941177}
\definecolor{color1}{rgb}{1,0.498039215686275,0.0549019607843137}
\definecolor{color2}{rgb}{0.172549019607843,0.627450980392157,0.172549019607843}

\begin{axis}[
width=5cm,
height=4cm,
legend cell align={left},
legend style={draw=white!80.0!black},
tick align=outside,
tick pos=left,
title={Pendulum},
title style={xshift=1mm, yshift=-1.5mm},
x grid style={white!69.01960784313725!black},
xlabel={Iterations},
xmin=-5, xmax=105,
xtick style={color=black},
y grid style={white!69.01960784313725!black},
ylabel={Cost},
ymin=11766.1516300143, ymax=41763.4969180331,
ytick style={color=black}
]

\addplot [semithick, color1, mark=*, mark size=1, mark options={solid}]
table {%
0 40399.9812231231
1 40399.9793276954
2 40399.9772136019
3 40399.9748488573
4 40399.9721957004
5 40399.9692093564
6 40399.9658364929
7 40399.9620132838
8 40399.9576629657
9 40399.9526927382
10 40399.9469898046
11 40399.9404162833
12 40399.9328026228
13 40399.9239390185
14 40399.9135641426
15 40399.9013502281
16 40399.8868831656
17 40399.8696357138
18 40399.8489311142
19 40399.8238931957
20 40399.7933772622
21 40399.7558733395
22 40399.7093692031
23 40399.6511541585
24 40399.5775344151
25 40399.4834147225
26 40399.3616747643
27 40399.2022257243
28 40398.9905603611
29 40398.7054871549
30 40398.3155260014
31 40397.773065729
32 40397.0047018787
33 40395.8949132849
34 40394.257853317
35 40391.7874147488
36 40387.9665513485
37 40381.8981194001
38 40371.9802886464
39 40355.2652964677
40 40326.155067603
41 40273.6740535328
42 40175.6431487553
43 39986.1627384451
44 39609.7069153871
45 38856.0963953669
46 37413.1069328776
47 35045.7888011232
48 32100.5616137964
49 29168.739582412
50 25361.602359586
51 23703.955799683
52 22952.3472204562
53 22325.9957790642
54 21845.656200462
55 21458.3075146997
56 21178.3573952902
57 20966.5169404051
58 20811.9672945773
59 20674.7514574354
60 20543.4689653907
61 20429.7292417623
62 20321.9779009981
63 20228.2385564261
64 20140.8241586201
65 20078.8653236962
66 20013.9937984607
67 19962.7336460591
68 19922.8623063177
69 19880.0196510702
70 19843.3959461837
71 19799.2870611319
72 19777.689515926
73 19763.636592026
74 19744.9935329011
75 19730.2465470274
76 19708.4520350339
77 19685.7735246687
78 19683.3143533938
79 19675.6680896078
80 19668.2518204913
81 19540.8706493507
82 19124.6942639187
83 17241.352990928
84 17158.0746835301
85 17138.2371018075
86 17057.6154263642
87 16946.1419111275
88 16922.3513087063
89 16843.8616211627
90 16796.3922241822
91 16790.1479887662
92 16745.3952244375
93 16710.7595521732
94 16682.0025157546
95 16662.8822889816
96 16653.3307792758
97 16634.2414452088
98 16629.3926208909
99 16601.4832405629
100 16598.7564748258
};

\addplot [semithick, color2, mark=*, mark size=1, mark options={solid}]
table {%
0 40399.9424911043
1 40399.0877730383
2 40396.4754077926
3 40391.2387362077
4 40379.1443142081
5 40335.0805410575
6 40165.2836109056
7 39903.744772495
8 39558.5378366876
9 39137.9906043485
10 38650.9652664155
11 38106.6702766302
12 37514.2919000055
13 36882.5679416221
14 36219.3502818842
15 35531.2855808465
16 34823.4510915338
17 34099.1392646206
18 33359.524514119
19 32603.3050868135
20 31826.594017765
21 31023.4203522275
22 30188.0043961603
23 29319.9176785315
24 28432.1631049755
25 27558.0433830204
26 26747.0578406579
27 26043.1688264505
28 25459.1495895001
29 24974.3594050347
30 24556.737343215
31 24183.9049119086
32 23855.9364389067
33 23562.7834472807
34 23280.675878985
35 23024.7301426022
36 22781.6127592776
37 22564.3618933772
38 22343.3519267721
39 22152.4069666288
40 21968.1233850423
41 21816.4941252711
42 21654.7190682856
43 21510.9190860907
44 21393.3719898799
45 21293.9902190338
46 21171.3247466165
47 21096.9849227356
48 21029.0936910306
49 20935.2252937415
50 20870.549057077
51 20801.0202591946
52 20743.3972130713
53 20689.7134319363
54 20651.5915697142
55 20593.893369656
56 20548.4528307127
57 20515.9539197223
58 20479.7823007178
59 20445.0690516348
60 20403.8923332154
61 20381.3231610148
62 20347.9710575531
63 20330.54777582
64 20318.3348260455
65 20307.6188189226
66 20282.6198030713
67 20268.162882195
68 20250.9669435203
69 20235.1476237161
70 20219.1175065557
71 20197.9394102723
72 20190.0051338048
73 20178.8523111742
74 20174.7440512676
75 20172.1860396669
76 20165.5430863349
77 20155.3488309084
78 20144.4767236317
79 20136.3829102378
80 20118.4318604357
81 20113.5606115799
82 20105.2738030317
83 20095.1715761715
84 20092.4205042997
85 20083.3723773583
86 20075.1434647519
87 20068.6988643316
88 20064.5901150242
89 20060.2172851861
90 20054.9633074956
91 20051.8429916738
92 20045.7723269358
93 20040.2575291655
94 20038.5196974405
95 20034.4611663031
96 20034.4357019326
97 20026.3652751555
98 20023.167730018
99 20022.7758136791
100 20018.2135181369
};

\addplot [semithick, color0, mark=*, mark size=1, mark options={solid}]
table {%
	0 40400
	1 40400
	2 40400
	3 40400
	4 40400
	5 40400
	6 40400
	7 40400
	8 40400
	9 40400
	10 40400
	11 40400
	12 40400
	13 40400
	14 40400
	15 40400
	16 40400
	17 40400
	18 40400
	19 40400
	20 40400
	21 40399.9999999999
	22 40399.9999999997
	23 40399.9999999988
	24 40399.999999996
	25 40399.9999999864
	26 40399.9999999538
	27 40399.9999998436
	28 40399.9999994727
	29 40399.9999982292
	30 40399.9999940774
	31 40399.9999802698
	32 40399.9999345286
	33 40399.9997835831
	34 40399.9992873613
	35 40399.9976622129
	36 40399.9923596103
	37 40399.975121872
	38 40399.9192900764
	39 40399.7391104246
	40 40399.159735214
	41 40397.3034910328
	42 40391.3787103481
	43 40372.5485283352
	44 40313.0560668851
	45 40127.2179580899
	46 39563.8021964674
	47 38024.629115844
	48 35034.1809971427
	49 31004.6465670239
	50 24316.875931948
	51 19514.7851235382
	52 17799.3835991956
	53 14394.4433079234
	54 14354.1247842105
	55 14276.8468443096
	56 14206.7658504559
	57 14137.3524684215
	58 14091.5095635071
	59 14011.7778377116
	60 13977.7368870451
	61 13939.9789560227
	62 13932.715790518
	63 13908.4239995746
	64 13876.7556580717
	65 13857.1645337621
	66 13839.9857257221
	67 13820.760720932
	68 13793.6927591319
	69 13740.2834266152
	70 13771.0397845351
	71 13757.6611370294
	72 13742.766197867
	73 13722.2144760438
	74 13710.0283904104
	75 13682.5420241333
	76 13694.8275711064
	77 13642.8432945481
	78 13645.0508340887
	79 13645.410303439
	80 13661.128328066
	81 13652.670505051
	82 13592.4771750152
	83 13632.7914143191
	84 13604.2791900885
	85 13613.1891618692
	86 13577.210809203
	87 13591.7614913816
	88 13577.4198021807
	89 13562.8497003964
	90 13524.8629255253
	91 13574.8800329039
	92 13556.4923021759
	93 13485.2999438683
	94 13550.0942614795
	95 13554.7922447622
	96 13522.5343547924
	97 13488.6309139673
	98 13529.4012541208
	99 13536.1836188076
};

\end{axis}

\end{tikzpicture}
	\end{minipage}
	\hfill
	\begin{minipage}[b][3.7cm][t]{0.32\textwidth}
		\centering
		\input{fig/Cartpole.tex}
	\end{minipage}
	\hfill
	\begin{minipage}[b][3.7cm][t]{0.32\textwidth}
		\centering
		\input{fig/DoubleCartpole.tex}
	\end{minipage}
	\hfill
	\begin{minipage}[t][.7cm][t]{\textwidth}
		\caption{Comparison of the trajectory cost prediction over iterations for three simulated tasks during trajectory optimization. 
		For all algorithms, the dynamics are linearized once per iteration. For experimental details see Section \ref{subsec:nonlinear_eval}.}\label{fig:known_cost}
	\end{minipage}
\end{figure}

\subsection{Evaluation on nonlinear trajectory optimization tasks}
\label{subsec:nonlinear_task}}
To evaluate the viability of \itwoc{} for nonlinear trajectory optimization, its performance on three standard control tasks were compared to similar baseline methods.
iLQR and GPS are two popular algorithms that use local linearization for time-varying controllers and have demonstrated strong performance on complex control problems. 
iLQR is deterministic, so here it is used as a baseline for the ignoring uncertainty in stochastic control problems.
While GPS was motivated to train Neural Network policies, here we use its time-varying linear controllers, viewing it as Maximum Entropy iLQG.
In order to perform the linearization required for approximate inference (and the baseline approaches), the test environments were implemented using the \texttt{Autograd} library \cite{maclaurin2015autograd}.
We test on three classical problems of increasing complexity in state-action-observation dimensionality ($d_x$, $d_u$, $d_z$): Pendulum (2, 1, 4), Cartpole (4, 1, 6) and Double Cartpole (6, 1, 9) swing-up.
Both Cartpole domains are also underactuated, which presents a significant planning challenge. 
All environments also have constrained actuation, which introduces both a nonlinearity and increased sensitivity to disturbances.
Experimental details and additional trajectory plots are included in  Section~\ref{subsec:nonlinear_experiment_details}.

\begin{figure}[b!]
	\begin{minipage}[t]{\columnwidth}
		\centering
		\begin{tikzpicture}

\definecolor{color0}{rgb}{0.12156862745098,0.466666666666667,0.705882352941177}
\definecolor{color1}{rgb}{1,0.498039215686275,0.0549019607843137}
\definecolor{color2}{rgb}{0.172549019607843,0.627450980392157,0.172549019607843}

\begin{axis}[
hide axis,
width=14cm,
xmin=10, xmax=50,
ymin=0, ymax=1.0,
legend cell align={center},
legend columns=3,
legend style={/tikz/every even column/.append style={column sep=1.5cm}}]
]

\addlegendimage{no markers, line width=1.0pt, color0}
\addlegendentry{\small I2C};
\addlegendimage{no markers, line width=1.0pt, color1}
\addlegendentry{\small iLQR };
\addlegendimage{no markers, line width=1.0pt, color2}
\addlegendentry{\small GPS};

\end{axis}

\end{tikzpicture}
	\end{minipage}
	\hfill
	\begin{minipage}[b]{\columnwidth}
		\input{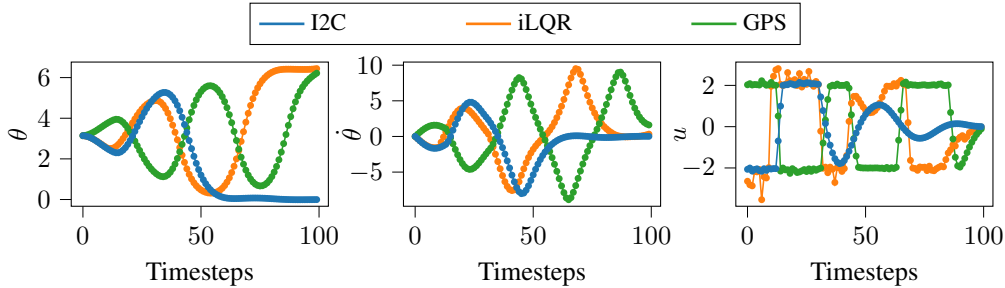}
		\caption{Comparison of the state-action trajectories of \itwoc{}, iLQR and GPS on the Pendulum swing-up task after convergence.}
		\label{fig:pendulum traj}
	\end{minipage}
	\vspace{-9mm}
\end{figure}

Figure \ref{fig:known_cost} shows that \itwoc{} is capable of performing effective trajectory optimization.
The EM aspect of the algorithm results in a significant portion of the time is used `warming up' the priors, which are set to be small in order to carry out steady exploration, rather than optimizing the control cost.
iLQR performs superior trajectory optimization, both in rate and final cost.
However, actuation constraints were found to lead to suboptimal convergence (in the Pendulum task, Figure \ref{fig:pendulum traj}), and the optimized controllers were comparatively highly aggressive.
GPS performed steadier optimization due to the KL bound and exploration in the forward pass.
In Table \ref{tab:ctrl_eval}, the optimized (deterministic) controllers were evaluated on the stochastic environment.
\itwoc{} performs the most consistently, operating close to its predicted cost for each task. 
GPS and iLQR, with more aggressive controllers and trajectories, both suffered reduced performance when evaluated on the simulated systems.
We attribute this to the high-risk strategy of operating at the actuation limits, when also subjected to disturbances, especially as time-varying control strategies are inherently very brittle to any deviation in trajectory.
{\renewcommand{\arraystretch}{1}
	\begin{table}[t]
		\centering
		\vspace{-3mm}
		\begin{tabular}{llll}
			\toprule
			Environment & Algorithm  & Predicted Cost & Evaluated Cost\\
			\midrule
			\multirow{3}{*}{Pendulum}        & \itwoc{} & $\bm{1.35\sn{4}}$        & $\bm{1.37\sn{4}\pm3.82}$ \\
			& iLQR     & $1.66\sn{3}$ & $1.11\sn{5}\pm20.38$     \\
			& GPS      & 2.00\sn{4}        & $7.01\sn{4}\pm30.96$     \\
			\midrule	                                 		
			\multirow{3}{*}{Cartpole}        & \itwoc{} & 1.73\sn{5}        & $\bm{1.74\sn{5}\pm0.14}$ \\
			& iLQR     & $\bm{1.14\sn{5}}$ & $1.76\sn{7}\pm88.63$ \\
			& GPS      & 1.65\sn{5}        & $2.94\sn{6}\pm17.60$ \\
			\midrule	
			\multirow{3}{*}{Double Cartpole} & \itwoc{} & 3.12\sn{5}        & $\bm{3.21\sn{5}\pm1.79}$  \\
			& iLQR     & $\bm{2.37\sn{5}}$ & $1.76\sn{7}\pm5.27\sn{5} $  \\
			& GPS      & 3.76\sn{5}        &  $2.94\sn{6}\pm44.39$ \\
			\bottomrule\vspace{0mm}
		\end{tabular}
		\caption{Evaluating the optimized deterministic controller of each algorithm on the simulated stochastic environments.
			Predicted Cost refers to the converged value from Figure \ref{fig:known_cost},
			Evaluated Cost shows the mean and standard deviation after 100 trials.}
		\label{tab:ctrl_eval}
		\vspace{-7mm}
	\end{table}
}
\section{Related Work}
\label{sec:related}
Optimal control of nonlinear dynamical systems through iterative linearization originated from Differential Dynamic Programming (DDP) \cite{jacobson1970differential}.
A drawback of DDP is the need for the computationally expensive second-order approximation of the dynamics.
In the framework of Iterative LQR (iLQR) \cite{DBLP:conf/icinco/LiT04} and its stochastic extension iLQG \cite{todorov2005generalized}, this requirement is dropped.
Both algorithms perform only first order approximations, making them akin to a regularized Gauss-Newton method.
All former methods however lack a principled forward pass and instead rely on a line-search approach to find a suitable regularization, that counteracts the greediness of their local approximations.
Extended LQR (eLQR) \cite{van2016extended} and its stochastic extension seLQR \cite{DBLP:journals/tase/SunBA16} address this issue and perform a forward pass based on the `cost-to-come', that has similarities to Kalman filtering.
A more elegant solution to the problem of regularization is proposed in Guided Policy Search (GPS) \cite{levine2014motor,levine2014learning}, where the Stochastic Optimal Control problem is formulated with a KL bound on the change of trajectory distributions.
GPS derives a Maximum Entropy iLQG as a means to train neural network policies.

The connection between optimal control and inference, also known as the estimation-control duality and Kalman duality \cite{stengel1986stochastic, todorov2008general} was initially noted by Kalman \cite{kalman1960new}, while working on the Kalman Filter and Optimal Control.
Probabilistic Control Design \cite{karny1996towards, karny2006fully, vsindelavr2008stochastic} derives a probabilistic variant of LQR through a KL divergence minimization, also noting the connection between the LQR cost weight matrices and the precisions of multivariate normal distributions.
Furthermore, the similarity between LQR and Kalman Smoothed trajectories has previously been utilized for the ERTS controller \cite{zima2013extended}.
However, this work uses standard smoothing in the state and does not derive a corresponding controller, relying on an approximate inverse dynamics model instead.

Inference has been applied to reinforcement learning for discrete environments \cite{Attias03planningby}, through maximizing the likelihood of a discrete latent optimality variable.
AICO \cite{toussaint2009robot} applies this approach to the continuous LQR setting, with the state cost defining the optimality probability, and the action weight defining the precision of the action prior, which is treated like a disturbance for exploration.
As with \itwoc{}, the backward messages were found to share similarities with the DAREs of LQR, however unlike \itwoc, the input priors are fixed.
AICO was generalised to Posterior Policy Iteration (PPI) \cite{rawlik2013stochastic, rawlik2013probabilistic} in which a risk-tuned linear Gaussian controller was obtained from the inferred value function.
The idea of controls as a random diffusion process is shared by Todorov \cite{todorov2007linearly}, along with Path Integral (PI) Control (KL Control for discrete environments) \cite{DBLP:conf/aips/KappenGO13,DBLP:conf/nips/PanTK15,DBLP:conf/pkdd/GomezKPN14}
that takes advantage of Feynman-Kac lemma to approximately solve the continuous-time Hamilton-Jacobi-Bellman equation using stochastic processes.
PI methods iteratively compute local improvements to the controls, allowing them to be used to train parametric policies or for model predictive control.
\section{Conclusion}
\label{sec:conclusion}
In this work we have introduced Input Inference for Control (\itwoc{}), a novel control-as-inference formulation, by casting optimal control as Bayesian inference over the inputs.
Through making the linear Gaussian assumption, we arrived at a tractable approximate EM algorithm with the use of message passing for approximate inference, and are able to draw connections with linear quadratic optimal control, Kalman filtering and Kalman smoothing through examination of the messages.
Compared to prior work, this scheme employs natural regularization through the mechanisms of Bayesian inference, offering a more principled approach than currently established deterministic solvers.
Moreover, our approach improves previous probabilistic approaches by naturally incorporating and optimizing over actions, enabling us to retrieve time-variant feedback controllers.
Future avenues of research include the analysis of different approximate inference techniques, such as Monte Carlo, variational methods, and numerical quadrature, and the investigation of the trade-off between accuracy of inference, computational cost and benefit to control optimization.
\clearpage
\acknowledgments{
The authors would like to thank Michael Lutter and Julen Urain for valuable feedback on the draft.
This project has received funding from the European Union’s Horizon 2020 research and innovation programme under grant agreement No. 640554 (SKILLS4ROBOTS).
}

\bibliography{lib}
\newpage
\label{sec:app}
\appendix
\section{The Dynamic Programming solution to the Linear Quadratic Regulator}
\label{subsec:lqr}
Given a linear system, we wish to find a control sequence $\vu^*_{0:T}$ that minimizes a quadratic cost function over a finite time horizon $T$ for a goal state $\vx_g$ and input $\vu_g$:
\begin{align}
    &\min_{\vu_{0:T}}\left[(\vx_T\!-\!\vx_g)\tran \mQ_f (\vx_T\!-\!\vx_g) + 
    \textstyle\sum^{T-1}_{t=0}(\vx_t\!-\!\vx_g)\tran \mQ (\vx_t\!-\!\vx_g) + (\vu_t\!-\!\vu_g)\tran\mR(\vu_t\!-\!\vu_g)\right] \notag\\
    &\text{s.t.}\;\vx_{t+1} = \mA\vx_t + \va + \mB \vu_t \label{eq:lqr}
\end{align}
Solving this method via Dynamic Programming, we can construct a quadratic value function backwards through time to find the optimal control at each timestep, which we can calculate for  \Eqref{eq:lqr} using Bellman's Principle of Optimality.
\begin{align}
\shortintertext{Starting with $\mP_T = \mQ_f,\;\vp_T = \bm{- \vx_g {\tran} \mQ},\;p_T = 0$.}
V_t(\vx) &= \vx\tran\mP_t\vx + 2\vx\tran\vp_t + p_{t} \label{eq:value_func}\\
         &= \min_{\vu}\left[(\vx_t-\vx_g)\tran\mQ(\vx_t-\vx_g) + (\vu_t-\vu_g)\tran\mR(\vu_t-\vu_g) + V_{t+1}(\vx_{t+1})\right]
\intertext{The optimal input can be found to be linear in state,}
    \vu^*_t &= -(\mR + \mB\tran\mP_{t+1}\mB)\inv(\mB\tran\mP_{t+1}(\mA\vx_t + \va) + \mB\tran\vp_{t+1} - \mR\vu_g) \label{eq:u_lqr} \\
            &= \mK_t \vx_t + \vk_t
\intertext{The parameters of the value functions follow the recursive form, i.e.}
    \mP_t &= \mQ + \mA\tran \mP_{t+1} \mA + \mA\tran \mP_{t+1} \mB (\mR + \mB\tran \mP_{t+1} \mB)\inv\mB\tran\mP_{t+1}\mA\label{eq:P_lqr}\\
    \vp_{t} &= \mA\tran(\mP_{t+1}\va \!+\!\vp_{t+1}\!-\!\mP_{t+1}\mB(\mR\!+\!\mB\tran\mP_{t+1}\mB)\inv(\mB\tran\mP_{t+1}\va + \mB\tran\vp_{t+1}\!-\!\mR\vu_g)) \notag \\
    &\hspace{5mm}\!-\!\mQ\vx_g \label{eq:p_lqr}
\end{align}
\section{Derivation of \textsc{i2c} Linear Gaussian Messages}
\label{subsec:msg}
All messages are derived following the graphical model in Figure~\ref{fig:FactorGraph}. Note the figure includes the intermediate variables (denoted with primes), used to add clarity to the derivations.
\subsection{Forward Messages}
\label{subsec:forward_msg}
The forward message are very close to those of Kalman Filtering, except the inputs are also observed and so have their own innovation step.
\begin{align}
\shortintertext{The innovation and propagation of the input into the system dynamics:}
    \nur{u}{'}{t} &= \nur{u}{~}{t} + \mF_t\tran (\sigXi + \mE_t\sigr{x}{~}{t}\mE_t\tran)\inv(\vz_t - \mE_t\mur{x}{~}{t} - \ve_t) \label{eq:fwd_start}\\
    \lamr{u}{'}{t} &= \lamr{u}{~}{t} + \mF_t\tran (\sigXi + \mE_t\sigr{x}{~}{t}\mE_t\tran)\inv \mF_t \\
    \mur{u}{''}{t} &= \mB_t\mur{u}{'}{t} \\
    \sigr{u}{''}{t} &= \mB_t\sigr{u}{'}{t}\mB_t\tran\label{eq:u2f} \\
\shortintertext{The innovation and propagation of the state, incorporating the input:}
    \nur{x}{'}{t} &= \nur{x}{~}{t} + \mE_t\tran(\sigXi + \mF_t\sigr{u}{~}{t}\mF_t\tran)\inv (\vz_t - \mF_t\mur{u}{~}{t} - \ve_t) \hspace{3.8cm} \\  
    \lamr{x}{'}{t} &= \lamr{x}{~}{t} + \mE_t\tran (\sigXi + \mF_t\sigr{u}{~}{t}\mF_t\tran)\inv \mE_t \\
    \mur{x}{''}{t} &=  \mA_t\mur{x}{'}{t} + \va_t \\
    \sigr{x}{''}{t} &=  \mA_t\sigr{x}{'}{t}\mA_t\tran \\
	\mur{x}{'''}{t} &=  \mur{x}{''}{t} \\
	\sigr{x}{'''}{t} &= \sigr{x}{''}{t} + \sigEta{t} \\
	\mur{x}{~}{t+1} &=  \mur{x}{'''}{t} + \mur{u}{''}{t} \\
	\sigr{x}{~}{t+1} &=  \sigr{x}{'''}{t} + \sigr{u}{''}{t} \label{eq:fwd_end}
\end{align}
\subsection{Backward Messages}
\label{subsec:backward_msg}
The most efficient means of constructing the backward messages for marginalisation is to make use of the `auxiliary' form (see \cite{loeliger2007factor, bruderer2015input}), which has several useful properties for message propagation. In particular they are invariant to the Addition factor, so the various offsets are automatically considered.  
\begin{align}
    \lama{x}{~}{~} &= (\sigr{x}{~}{~} + \sigl{x}{~}{~})\inv = 
    \lamr{x}{~}{~} - \lamr{x}{~}{~}\sigm{x}{~}{~} \lamr{x}{~}{~} \\
    \nua{x}{~}{~} &= \lama{x}{~}{~}(\mur{x}{~}{t} - \mul{x}{~}{t}) = \nur{x}{~}{t} - \lamr{x}{~}{t}\mum{x}{~}{t}
\end{align}
Like the marginal they are a fusion of the forward and backward message, but in the `dual' form.

For initialising the backward pass there are two approaches. One is to follow the idea of the terminal cost from \Eqref{eq:lqr}, where for example $\mP_T = \mQ_f = \mQ, \vp_T = \bm{0}$, $\laml{x}{~}{T} = \lamXi, \nul{x}{~}{T}=\bm{0}$. The marginals can then be constructed following \Eqref{eq:marginal}. Any $\mQ_f$ can be used, so long as $\laml{x}{~}{T}$ is constructed with an $\alpha$ following Section \ref{subsec:linear_gaussian} .
In practice, it was found crucial to tune up this terminal cost to ensure the target state is reached with a responsive controller (as many target states lay at unstable equilibria). However $\mQ_f$ the represents another (multi-dimensional) hyperparameter to tune. Using the probabilistic perspective, instead choose $\sigm{x}{~}{T}$ such that the prior $\sigr{x}{~}{T}$ has been reduced by a scale factor $\kappa$. While this deviates from the previous quadratic cost formulation into an adaptive cost function, it was found to be both simple and effective when tackling difficult domains. The adaptation becomes an important quality for nonlinear problems where the initial dynamics are stable and the target state dynamics are unstable. This scheme acts to tune up the terminal cost as the dynamics become more unstable, which causes the state uncertainty to grow at a greater rate, which in turn acts to increase the responsiveness of the controller. As we also wish to keep the prior and posterior trajectories tight during optimization (to ensure the linearization assumption is valid), we set $\mum{x}{~}{T}=\mur{x}{~}{T}$.
In the experiments of Section \ref{subsec:nonlinear_task}, $\mQ_f{=}\mQ$.
\begin{align}
\shortintertext{Starting with $\sigm{x}{}{T}=\sigr{x}{}{T},\;\mum{x}{}{T}=\mur{x}{}{T}$.}
\shortintertext{Construct the auxillary for $\vx_{t+1}$,}
    \lama{x}{~}{t+1} &= \lamr{x}{~}{t+1} - \lamr{x}{~}{t+1}\sigm{x}{~}{t+1}\lamr{x}{~}{t+1}\hspace{7.3cm} \label{eq:bwd_start}\\
    \nua{x}{~}{t+1} &= \nur{x}{~}{t+1} - \lamr{x}{~}{t+1}\mum{x}{~}{t+1} \\
\shortintertext{The auxiliary is invariant across an addition operation, so}
    \lama{x}{''}{t} &= \lama{x}{'''}{t} = \lama{x}{~}{t+1} \\
    \nua{x}{''}{t} &= \nua{x}{'''}{t} = \nua{x}{~}{t+1} \\
\shortintertext{Propagate the state belief backwards through system dynamics,}
    \lama{x}{'}{t} &= \mA_t\tran\lama{x}{''}{t}\mA_t \\
    \nua{x}{'}{t} &= \mA_t\tran\nua{x}{''}{t} \\
\shortintertext{Marginalized variables are invariant across the Equality node, so marginalize $\vx_t$ at $\vx_t'$,}
    \sigm{x}{}{t} &= \sigm{x}{'}{t} = \sigr{x}{'}{t} - \sigr{x}{'}{t}\lama{x}{'}{t}\sigr{x}{'}{t} \\
    \mum{x}{}{t} &= \mum{x}{'}{t} = \mur{x}{'}{t} - \sigr{x}{'}{t}\nua{x}{'}{t} \\
\shortintertext{To find $\mum{u}{~}{t}$, note that due to the addition operation, the auxillary of $\vu_t''$ is equal to that of $\vx_t'''$,}
    \lama{u}{''}{t} &= \lama{x}{'''}{t} \\
    \nua{u}{''}{t} &= \nua{x}{'''}{t} \\
    \lama{u}{'}{t} &= \mB_t\tran\lama{u}{''}{t}\mB_t\ \\
    \nua{u}{'}{t} &= \mB_t\tran\nua{u}{''}{t} \\
    \sigm{u}{~}{t} &= \sigm{u}{'}{t} = \sigr{u}{'}{t} -\sigr{u}{'}{t}\lama{u}{'}{t}\sigr{u}{'}{t} \\
    \mum{u}{~}{t} &= \mum{u}{'}{t} = \mur{u}{'}{t} - \sigr{u}{'}{t}\nua{u}{'}{t} \label{eq:bwk_end}
\end{align}
\subsection{Riccati Backward Messages for Control}
\label{subsec:ric_msg}
To understand the relation to optimal control, the backward messages must be represented recursively as a Discrete Algebraic Ricatti Equation.
\begin{align}
\shortintertext{\textbf{Recursion of the precision}}
    \laml{x}{}{t} &= \laml{x}{'}{t} + \mE_t\tran\laml{z}{'}{t}\mE_t\\
    &= \mA_t\tran\laml{x}{''}{t}\mA_t + \mE_t\tran\laml{z}{'}{t}\mE_t \\
    \laml{z}{'}{t} &= (\sigXi + \mF_t\sigr{u}{~}{t}\mF_t\tran)\inv
\shortintertext{\hspace{0cm}{Using the matrix inversion identity $(\mA\inv + \mB)\inv = \mA - \mA(\mA + \mB\inv)\mA $}\cite{petersen2008matrix},}
    \laml{x}{''}{t} &= (\sigEta{t} + \sigr{u}{''}{t} + \sigl{x}{~}{t+1})\inv\\
    &= \laml{x}{~}{t+1} - \laml{x}{~}{t+1}
        ((\sigEta{t} + \sigr{u}{''}{t})\inv + \laml{x}{~}{t+1})\inv\laml{x}{~}{t+1}\label{eq:lam2b}\\
    \sigr{u}{''}{t} &=  \mB_t(\lamr{u}{~}{t} + \mF_t\tran(\sigXi + \mE_t\sigr{x}{~}{t}\mE_t\tran)\inv\mF_t)\inv\mB_t\tran
\shortintertext{\hspace{0cm}{So the recursion in full is}}
    \laml{x}{~}{t} &= \mE_t\tran(\sigXi\!+\!\mF_t\sigr{u}{~}{t}\mF_t\tran)\inv\mE_t\!+\!\mA_t\tran\laml{x}{~}{t+1}\mA_t -\mA_t\tran\laml{x}{~}{t+1}\notag\\ 
    &\hspace{0.5cm}((\sigEta{t}\!+\!\mB_t(\lamr{u}{~}{t}\!+\! \mF_t\tran(\sigXi\!+\! \mE_t\sigr{x}{~}{t}\mE_t\tran)\inv\mF_t)\inv\mB_t\tran)\inv \!+\!\laml{x}{~}{t+1})\inv\laml{x}{~}{t+1}\mA_t\label{eq:lam_ric}
\shortintertext{\textbf{Recursion of the scaled-mean}}
    \nul{x}{~}{t} &= \nul{x}{'}{t} + \mE_t\tran\laml{z}{'}{t}(\vz_t-\mF_t\mur{u}{~}{t} -\ve_t)  \\
    \nul{x}{'}{t} &= \mA_{t}\tran\laml{x}{''}{t}(\mul{x}{''}{t} - \va_{t}) =
    \mA_{t}\tran\laml{x}{''}{t}(\sigl{x}{~}{t+1}\nul{x}{~}{t+1} - \mur{u}{''}{t} - \va_{t})
    \label{eq:x1b}
\shortintertext{{Substituting \Eqref{eq:fwd_start}-\ref{eq:u2f} into \Eqref{eq:x1b}} (using $\laml{z}{''}{t}$ for brevity)}
    \nul{x}{'}{t} &= \mA_{t}\tran\laml{x}{''}{t}(\sigl{x}{~}{t+1}\nul{x}{~}{t+1}\!-\!\va_{t}\!-\!\mB_{t}\sigr{u}{'}{t}(\nur{u}{~}{t}+\mF_t\tran\laml{z}{''}{t}(\vz_{t}-\mE_t\mur{x}{~}{t}-\ve_t)))
\shortintertext{{Substituting $\laml{x}{''}{t}$ through \Eqref{eq:lam2b}},}
     \nul{x}{'}{t} &= \mA_{t}\tran(\laml{x}{~}{t+1}\!-\!\laml{x}{~}{t+1}
        ((\sigEta{t} + \sigr{u}{''}{t})\inv\!+\! \laml{x}{~}{t+1})\inv\laml{x}{~}{t+1})\notag\\
        &\hspace{.8cm}(\sigl{x}{~}{t+1}\nul{x}{~}{t+1}\!-\!\va_{t}\!-\! (\mB_{t}(\lamr{u}{~}{t} + \mF_t\tran\laml{z}{''}{t}\mF_t)\inv(\nur{u}{~}{t}+\mF_t\tran\laml{z}{''}{t} \vz_{t})))   
\intertext{So the full recursion is,}
     \nul{x}{~}{t} &= \mA_{t}\tran(\mI\!-\!\laml{x}{~}{t+1}
        ((\sigEta{t} + \mB_t(\lamr{u}{~}{t} + \mF_t\tran(\sigXi + \mE_t\sigr{x}{~}{t}\mE_t\tran)\inv\mF_t)\inv\mB_t\tran)\inv\!+\! \laml{x}{~}{t+1})\inv)\notag\\
        &\hspace{.8cm}(\nul{x}{~}{t+1}\!-\!\laml{x}{~}{t+1}\va_{t}-\!\laml{x}{~}{t+1}(\mB_{t}(\lamr{u}{~}{t} + \mF_t\tran(\sigXi + \mE_t\sigr{x}{~}{t}\mE_t\tran)\inv\mF_t)\inv\notag\\
        &\hspace{.8cm}(\nur{u}{~}{t}+\mF_t\tran(\sigXi + \mE_t\sigr{x}{~}{t}\mE_t\tran)\inv(\vz_{t}-\mE_t\mur{x}{~}{t}-\ve_t))))\notag\\
    &\hspace{.4cm}+\!\mE_t\tran(\sigXi + \mF_t\sigr{u}{~}{t}\mF_t\tran)\inv(\vz_t-\mF_t\mur{u}{~}{t} -\ve_t)\label{eq:nu_ric} 
\end{align}
\subsection{Linear Gaussian Controller}
\label{subsec:controller_msg}
To extract the linear Gaussian controllers, we find the conditional distribution between $\vu_t$ and $\vx_t$. 
\begin{align}
\intertext{{The input estimate is marginalized by fusing the forward and backward message:}}
    \mum{u}{~}{t} &= \sigm{u}{~}{t}(\nur{u}{~}{t} + \nul{u}{~}{t}) \\
    \nul{u}{~}{t} &= \mF_t\tran(\sigXi+\mE_t\tran\sigr{x}{}{t}\mE_t)\inv(\vz_t-\mE_t\mur{x}{~}{t}-\ve_t) + \mB_t\tran\nul{u}{''}{t} \\
                  &= \mF_t\tran(\sigXi\!+\!\mE_t\tran\sigr{x}{}{t}\mE_t)^{\text{-}1}(\vz_t\!-\!\mE_t\mur{x}{~}{t}\!-\!\ve_t) + \mB_t\tran\laml{u}{''}{t}\mul{u}{''}{t} \\
                  &=  \mF_t\tran(\sigXi+\mE_t\tran\sigr{x}{}{t}\mE_t)^{\text{-}1}(\vz_t-\mE_t\mur{x}{~}{t}-\ve_t) \notag\\
                  &\hspace{5mm}+\mB_t\tran\laml{u}{''}{t}(\mul{x}{~}{t+1}\!-\!\mur{x}{'''}{t})\label{eq:start_of_pain} \\
\intertext{Eventually we need an expression in terms of the marginal $\vx_t$, so we need to be able to express Eq.~(\ref{eq:start_of_pain}) in terms of $\mum{x}{}{t}$. Taking the marginalisation rule from \Eqref{eq:marginal},}
    \mum{x}{'''}{t} &= \mA\mum{x}{~}{t} + \va_t = \sigm{x}{'''}{t}(\nur{x}{'''}{t} + \nul{x}{'''}{t})\\
    \mur{x}{'''}{t} &= \sigr{x}{'''}{t}(\lamm{x}{'''}{t}\mum{x}{'''}{t} - \nul{x}{'''}{t})
    = \sigr{x}{'''}{t}(\lamm{x}{'''}{t}\mum{x}{'''}{t} - \nul{x}{'''}{t})\\
    \mul{x}{~}{t+1}\!-\!\mur{x}{'''}{t} &= \mul{x}{~}{t+1}\!+\!\sigr{x}{'''}{t}\nul{x}{'''}{t}-\sigr{x}{'''}{t}\lamm{x}{'''}{t}\mum{x}{'''}{t}
\intertext{\hspace{0cm}{To find $\laml{u}{''}{t}$, we can use the matrix inversion identity $(\mA\inv +\mB\inv)\inv\!=\!\mB(\mA\!+\!\mB)\inv\mA\!=\!\mA(\mA + \mB)\inv\mB $} \cite{petersen2008matrix}}
    \laml{u}{''}{t} &= (\sigl{x}{~}{t+1} + \sigr{x}{'''}{t})\inv \\ 
                    &= \lamr{x}{'''}{t}(\laml{x}{~}{t+1} + \lamr{x}{'''}{t})\inv \laml{x}{~}{t+1} \\
    &= \laml{x}{~}{t+1}(\laml{x}{~}{t+1} + \lamr{x}{'''}{t})\inv\lamr{x}{'''}{t}\\
\shortintertext{We introduce dimensionless term $\mGamma$ for brevity, and will discuss interpretions of it in subsequent sections,}
	\mGamma_{t+1} &= \lamr{x}{'''}{t}(\laml{x}{~}{t+1} + \lamr{x}{'''}{t})\inv\\
	\mI - \mGamma_{t+1} &=  \laml{x}{~}{t+1}(\laml{x}{~}{t+1} + \lamr{x}{'''}{t})\inv\\
\shortintertext{This allows us to express the last term of \Eqref{eq:start_of_pain} as,}
\laml{u}{''}{t}(\mul{x}{~}{t+1}\!-\!\mur{x}{'''}{t}) &= (\mGamma_{t+1}\nul{x}{~}{t+1}\!+\!(\mI\!-\!\mGamma_{t+1})\nul{x}{'''}{t}\!-\!\mGamma_{t+1}\laml{x}{~}{t+1}\sigr{x}{'''}{t}\lamm{x}{'''}{t}\mum{x}{'''}{t}\!)\label{eq:mid-pain}\\
\shortintertext{where}
\nul{x}{'''}{t} &= \laml{x}{'''}{t}(\mul{x}{~}{t+1} - \mur{u}{''}{t})\label{eq:nulx2b}
\shortintertext{To develop the last term of \Eqref{eq:mid-pain}, recall the marginalisation rule for $\mLambda$ (\Eqref{eq:marginal}),}
\sigr{x}{'''}{t}\lamm{x}{'''}{t}\mum{x}{'''}{t} &= \sigr{x}{'''}{t}(\lamr{x}{'''}{t} + \laml{x}{'''}{t})\mum{x}{'''}{t}\label{eq:marg_x2}
\shortintertext{To understand this better, it is best to expand $\laml{x}{'''}{t}$,}
\laml{x}{'''}{t} &= (\sigl{x}{~}{t+1} + \sigr{u}{''}{t})\inv = \lamr{u}{''}{t}(\laml{x}{~}{t+1} + \lamr{u}{''}{t})\inv\laml{x}{~}{t+1}
\shortintertext{Applying this to \Eqref{eq:marg_x2} and introducing $\mPsi$ (another dimensionless scaling term)}
\sigr{x}{'''}{t}(\lamr{x}{'''}{t}\!+\!\laml{x}{'''}{t})\mum{x}{'''}{t}\!&=\sigr{x}{'''}{t}(\lamr{x}{'''}{t}\!+\!\lamr{u}{''}{t}(\laml{x}{~}{t+1}\!+\! \lamr{u}{''}{t})\inv\laml{x}{~}{t+1})\mum{x}{'''}{t} \\
&=\mPsi_{t+1}\mum{x}{'''}{t} \\
\text{where}\,\,\mPsi_{t+1} &= \sigr{x}{'''}{t}(\lamr{x}{'''}{t}\!+\!\lamr{u}{''}{t}(\laml{x}{~}{t+1}\!+\! \lamr{u}{''}{t})\inv\laml{x}{~}{t+1}) \label{eq:psi}
\shortintertext{To summarize:}
\mum{u}{~}{t} &= \sigm{u}{~}{t}(\nur{u}{~}{t} + \mF_t\tran(\sigXi\!+\!\mE_t\tran\sigr{x}{}{t}\mE_t)^{\text{-}1}(\vz_t\!-\!\mE_t\mur{x}{~}{t}\!-\!\ve_t)\notag\\
&\hspace{5mm}+\mB_t\tran\!(\mGamma_{t+1}\nul{x}{~}{t+1}\!+\!(\mI\!-\!\mGamma_{t+1}\!)\nul{x}{'''}{t} \notag\\
&\hspace{5mm}\!-\!\mGamma_{t+1}\laml{x}{~}{t+1}\mPsi_{t+1}(\!\mA\mum{x}{~}{t}\!+\!\va)))
\intertext{{To find $\sigm{u}{~}{t}$,}}
    \sigm{u}{~}{t} &= (\lamr{u}{~}{t} + \laml{u}{~}{t})\inv\\
    \sigm{u}{~}{t} &= (\lamr{u}{~}{t} + \mF_t\tran\lamXi\mF_t + \mB_t\tran(\sigl{x}{~}{t+1} + \sigr{x}{'''}{t})\inv\mB_t)\inv\\
\intertext{{Using the matrix inversion identity $(\mA\inv + \mB\inv)\inv = \mB(\mA + \mB)\inv\mA $} \cite{petersen2008matrix},}
    \sigm{u}{~}{t} &= (\lamr{u}{~}{t}\!+\!\mF_t\tran(\sigXi\!+\!\mE_t\tran\sigr{x}{}{t}\mE_t)^{\text{-}1}\mF_t\notag\\
    &\hspace{1cm}+\mB_t\tran\lamr{x}{'''}{t}(\lamr{x}{'''}{t}\!+\!\laml{x}{~}{t+1})\inv\laml{x}{~}{t+1}\mB_t)\inv\\
\sigm{u}{~}{t} &= (\lamr{u}{~}{t} + \mF_t\tran(\sigXi\!+\!\mE_t\tran\sigr{x}{}{t}\mE_t)\inv\mF_t + \mB_t\tran\mGamma_{t+1}\laml{x}{~}{t+1}\mB_t)\inv
\end{align}

\textbf{Interpreting the scale matrices $\mGamma$ and $\mPsi$}

Deriving the controller lead to the emergence of two scale matrices $\mGamma$ and $\mPsi$, representing matrix fractions of the forward messages (uncertainty) and backward messages (optimality). 
To interpret the meaning of these terms, there are four scenarios that are important to consider: high process uncertainty ($\lamr{x}{'''}{t}\rightarrow\bm{0}$), low process uncertainty ($\lamr{x}{'''}{t}\rightarrow\bm{\infty}$), high input prior uncertainty ($\lamr{u}{''}{t}\rightarrow\bm{0}$) and low input prior uncertainty ($\lamr{u}{''}{t}\rightarrow\bm{\infty}$). 
Note, `process uncertainty' includes accumulated uncertainty from previous timesteps, including that from the input priors. 
The input prior described above is specific to that timestep.
\begin{enumerate}
	\item \textbf{High process uncertainty, high input prior uncertainty} \\
	Here $\mGamma_{t+1}\rightarrow\bm{0}$, $\mPsi_{t+1}\rightarrow\bm{0}$ and $\laml{x}{'''}{t}\rightarrow\bm{0}$.
	Therefore the controller becomes cut off from the backward messages (i.e. any sense of optimality) and becomes a weighted average of it's prior and goal state.  
	\item \textbf{High process uncertainty, low input prior uncertainty} \\
	Here $\mGamma_{t+1}\rightarrow\bm{0}$, $\mPsi_{t+1}\rightarrow\mGamma_{t+1}\inv$ and $\laml{x}{'''}{t}\rightarrow\laml{x}{~}{t+1}$.
	Despite the system uncertainty, the controller confidence reactivates the control terms by cancelling out $\mGamma$.
	\item \textbf{Low process uncertainty, high input prior uncertainty} \\
	Here $\mGamma_{t+1}\rightarrow\mI$, $\mPsi_{t+1}\rightarrow\mI$ and $\laml{x}{'''}{t}\rightarrow\bm{0}$.
	This is the equivalent LQR setting, assuming the deterministic controller is used (see Section \ref{subsec:lqr_equiv}).
	\item \textbf{Low process uncertainty, low input prior uncertainty} \\
	Here $\mGamma_{t+1}\rightarrow\mI$, $\mPsi_{t+1}\rightarrow\mI$ and $\laml{x}{'''}{t}\rightarrow\laml{x}{~}{t+1}$.
	As above this is similar to the LQR setting, however now the controller update will be closer to its prior.
\end{enumerate}

\subsection{Equivalence to the Dynamic Programming LQR Solution}
\label{subsec:lqr_equiv}
First, remembering that the backwards message correspond to likelihoods, the log-likelihood of a Gaussian distribution is,
\begin{align}
    \log\mathcal{N}(\vx; \vmu, \mSigma) &= (\vx - \vmu)\tran\mSigma\inv(\vx - \vmu)+ \text{constant} \\
                                        &= \vx\tran\mSigma\inv\vx - 2\vx\tran\mSigma\inv\vmu + \text{constant} \\
                                        &= \vx\tran\mLambda\vx - 2\vx\tran\vnu + \text{constant}
\end{align}
By comparing this to the LQR value function in \Eqref{eq:value_func}, the equivalence between $\mLambda$, $\mP$, $-\vnu$ and $\vp$ outlined in Table \ref{tab:equiv} may be appreciated. 

To arrive at the recursive LQR expressions outlined in Section \ref{subsec:lqr} we must consider linear models, deterministic dynamics and infinitely broad priors, which requires $\sigm{v}{~}{~}\!\rightarrow\bm{0}$ and $\lamr{u}{~}{t}\!\rightarrow\bm{0}$.
Additionally, from the formulation outlined in Section \ref{subsec:linear_gaussian}, $\mE\tran\lamXi\mE=\alpha\mQ$ and $\mF_t\tran\lamXi\mF_t=\alpha\mR$.
To recover the LQR result, we require the observation likelihood to dominate, which occurs for suitable large $\alpha$.
Moreover, given large input priors, we require $\alpha$ such that
$(\sigXi + \mF
_t\sigr{u}{}{t}\mF_t\tran)\inv\approx\lamXi$, therefore $\alpha\rightarrow\infty$ as $\lamr{u}{~}{t}\!\rightarrow\bm{0}$.
As the value function parameters scale linearly with the cost function and the controller is invariant to the scale, we can omit $\alpha$ from the analysis for brevity.

\textbf{Recursion of the precision}
\begin{align}
\shortintertext{Applying these conditions to the $\laml{x}{~}{t}$ recursion in \Eqref{eq:lam_ric},}
    \laml{x}{~}{t} &= \mQ + \mA\tran\laml{x}{~}{t+1}\mA -\mA\tran\laml{x}{~}{t+1}((\mB_t\mR\inv\mB_t\tran)\inv + \laml{x}{~}{t+1})\inv\laml{x}{~}{t+1}\mA \label{eq:lam_equiv1}\\
    &= \mQ + \mA\tran\laml{x}{~}{t+1}\mA -\mA\tran\laml{x}{~}{t+1}\mB(\mR + \mB\tran\laml{x}{~}{t+1}\mB)\inv\mB\tran\laml{x}{~}{t+1}\mA \label{eq:lam_equiv2}
\end{align}
\vspace{-5mm}
\begin{align}
\shortintertext{\Eqref{eq:lam_equiv1} to \ref{eq:lam_equiv2} is achieved using the identity $(\mA + \mB)\inv = \mB\inv(\mB\inv + \mA\inv)\inv\mA\inv$\cite{petersen2008matrix},}
    ((\mB\mR\inv\mB\tran)\inv + \laml{x}{~}{t+1})\inv &= \sigl{x}{~}{t+1}(\mB\mR\inv\mB\tran + \sigl{x}{~}{t+1} )\inv\mB\mR\inv\mB\tran
\shortintertext{along with $ (\mA + \mJ\tran\mB\mJ)\inv\mJ\tran\mB = \mA\inv\mJ\tran(\mB\inv + \mJ\mA\mJ\tran)\inv$\cite{petersen2008matrix}}
    &= \mB(\mR + \mB\tran\laml{x}{~}{t+1}\mB)\inv\mB\tran
\end{align}

\textbf{Recursion of the scaled-mean}
\begin{align}
\shortintertext{Applying the conditions to the $\nul{x}{~}{t}$ recursion in \Eqref{eq:nu_ric}, along with the identity tricks used above,}
     \nul{x}{~}{t} &= \mA\tran(\mI\!-\!\laml{x}{~}{t+1}
    ((\mB\mR\inv\mB_t\tran)\inv\!+\! \laml{x}{~}{t+1})\inv)(\nul{x}{~}{t+1}\!-\!\laml{x}{~}{t+1}(\mB\mR\inv(\mR\vu_g)) - \laml{x}{~}{t+1}\va)\notag\\
    &\hspace{.4cm}+\!\mQ\vx_g \\
     &= \mA\tran(\mI\!-\!\laml{x}{~}{t+1}
    \mB(\mR\!+\! \mB\tran\laml{x}{~}{t+1}\mB\tran)\inv\mB\tran)(\nul{x}{~}{t+1}\!-\!\laml{x}{~}{t+1}(\mB\mR\inv(\mR\vu_g)) - \laml{x}{~}{t+1}\va)\notag\\
    &\hspace{.4cm}+\!\mQ\vx_g \\
    &= \mA\tran(\nul{x}{~}{t+1}-\laml{x}{~}{t+1}\va\!-\!\laml{x}{~}{t+1}\mB\vu_g \!-\!\laml{x}{~}{t+1}\mB(\mR\!+\!\mB\tran\laml{x}{~}{t+1}\mB\tran)\inv\mB\tran \notag\\
    &\hspace{.8cm}(\nul{x}{~}{t+1}\!-\!\laml{x}{~}{t+1}\mB\vu_g - \laml{x}{~}{t+1}\va)+\!\mQ\vx_g
\shortintertext{Here, there is a discrepancy in the $\vu_g$ terms, but this can be rectified through adding $\mR\vu_g-\mR\vu_g$ and rearranging,}
    &= \mA\tran(\nul{x}{~}{t+1}-\laml{x}{~}{t+1}\va\!-\!\laml{x}{~}{t+1}\mB\vu_g\!-\!\laml{x}{~}{t+1}\mB(\mR\!+\!\mB\tran\laml{x}{~}{t+1}\mB\tran)\inv \notag\\
    &\hspace{.8cm}(\mB\tran\nul{x}{~}{t+1}\!-\!\mB\tran\laml{x}{~}{t+1}\mB\vu_g + \mR\vu_g-\mR\vu_g - \mB\tran\laml{x}{~}{t+1}\va)+\!\mQ\vx_g \\
\shortintertext{$-(\mR\!+\!\mB\tran\laml{x}{~}{t+1}\mB\tran)\vu_g$ can be taken outside to cancel out the existing term there, so only one $\vu_g$ term remains,}
    &= \mA\tran(\nul{x}{~}{t+1}-\laml{x}{~}{t+1}\va\!-\!\laml{x}{~}{t+1}\mB\vu_g + \laml{x}{~}{t+1}\mB\vu_g\!-\!\laml{x}{~}{t+1}\mB(\mR\!+\!\mB\tran\laml{x}{~}{t+1}\mB\tran)\inv \notag\\
    &\hspace{.8cm}(\mB\tran\nul{x}{~}{t+1}\!+\!\mR\vu_g\!-\!\mB\tran\laml{x}{~}{t+1}\va)+\!\mQ\vx_g \\
    &= \mA\tran(\nul{x}{~}{t+1}-\laml{x}{~}{t+1}\va\!-\!\laml{x}{~}{t+1}\mB(\mR\!+\!\mB\tran\laml{x}{~}{t+1}\mB\tran)\inv \notag\\
    &\hspace{.8cm}(\mB\tran\nul{x}{~}{t+1}\!-\!\mB\tran\laml{x}{~}{t+1}\va\!+\!\mR\vu_g)+\!\mQ\vx_g
\end{align}
Recall that $\vp_t$ is equivalent to $-\vnu_t$, so all non-$\vnu$ terms should have the opposite sign to those in Eq.~\ref{eq:p_lqr}. 

\textbf{The linear Gaussian controller}

As mentioned above, for the LQR conditions $\mGamma_{t+1}\rightarrow\mI$, $\mPsi_{t+1}\rightarrow\mI$ and $\laml{x}{'''}{t}\rightarrow\bm{0}$.
\begin{align}
    \shortintertext{Applying the conditions to the controller,}
	\sigm{u}{~}{t} &= (\mR + \mB_t\tran\laml{x}{~}{t+1}\mB_t)\inv, \\
    \mum{u}{~}{t} &= -\sigm{u}{~}{t}(-\mR\vu_g+\mB_t\tran(-\nul{x}{~}{t+1}+\laml{x}{~}{t+1}(\mA\mum{x}{~}{t}+\va))),
\end{align}
remembering that $\nul{x}{~}{t+1}$ is the opposite sign to $\vp_{t+1}$.

Expanding on the case of highly uncertainty, where $\mGamma_t\rightarrow\bm{0}$, here the stochastic controller is independent of the backward messages (and therefore any notion of optimality). Therefore it would depend purely on a weighted combination of its prior and goal:
\begin{align}
    \sigm{u}{~}{t} &= (\lamr{u}{~}{t} + \alpha\mR)\inv, \\
    \mum{u}{~}{t} &= (\lamr{u}{~}{t} + \alpha\mR)\inv(\nur{u}{~}{t} + \alpha\mR\vu_g),
\end{align}
where the stationary distribution would be $\sigm{u}{~}{t}\rightarrow\bm{0}$ and $\mum{u}{~}{t}\rightarrow\bm{0}$. 
\begin{figure}[b]
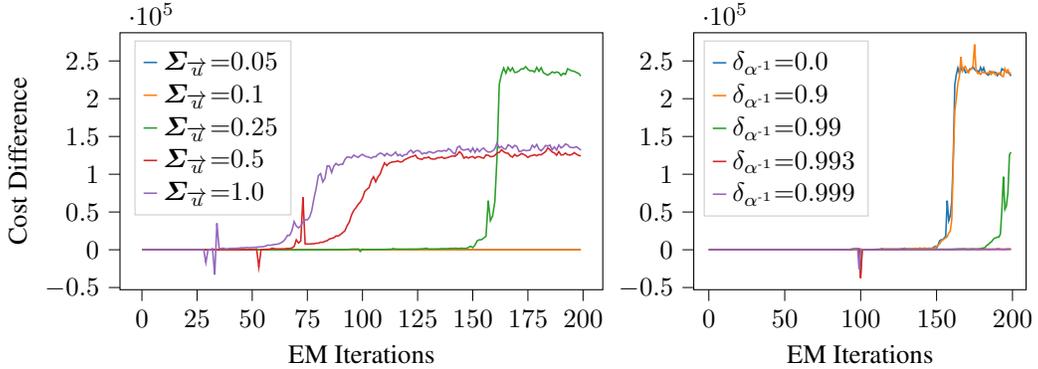

\begin{minipage}[b][5cm][t]{0.49\textwidth}
	\centering
	\input{fig/sig_u_sensitivity.tex}
\end{minipage}
	\hfill
\begin{minipage}[b][5cm][t]{0.49\textwidth}
	\hspace{9mm}
	\input{fig/alpha_sensitivity.tex}
\end{minipage}

\hfill
\begin{minipage}[t][.8cm][t]{0.55\textwidth}
	\centering
	\vspace{-.4cm}
	\subcaption{Cost difference over iterations with varying $\sigr{u}{}{}$ for $\delta_{\alpha\inv}{=}0$. Smaller values delay the point of divergence, although optimization progress is also slowed.}\label{fig:sig_u_sense}
\end{minipage}
\hfill
\begin{minipage}[t][.8cm][t]{0.4\textwidth}
	\centering
	\vspace{-.4cm}
	\subcaption{Cost difference over iterations with varying $\delta_{\alpha\inv}$ for $\sigr{u}{}{}{=}0.25$. Increasing $\delta_{\alpha\inv}$ stabilizes the approximate inference.}\label{fig:delta_alpha_sense}
\end{minipage}
\hfill
\begin{minipage}[t][.7cm][t]{\textwidth}
	\caption{Difference between evaluated cost and predicted cost over EM iterations across hyperparameters for the Cartpole swing-up task.}\label{fig:sensitivity}
\end{minipage}
\end{figure}
\subsection{Hyperparameter Sensitivity}
For nonlinear tasks, the crucial hyperparameters for \itwoc{} are the initial input priors $\sigr{u}{}{}$ and the update limit (motivated as a KL bound) of $\alpha$, $\delta_\alpha$.
The role of $\sigr{u}{}{t}$ is to facilitate exploration, but too much uncertainty in the trajectory leads to the linearization assumption becoming invalid during inference.
This failure mode manifests as the posterior inputs becoming inaccurate, therefore leading to the subsequent prior trajectory deviating from the previous posterior trajectory.
This means that the predicted performance of the controller diverges from the true performance when evaluated on the actual system.  
Therefore, for fast and successful convergence, $\sigr{u}{}{}$ depends not only on the expected input range, but should also be tuned based on the inherent uncertainty of the system and nonlinearity of the dynamics.

Even after tuning $\sigr{u}{}{}$, the approximate inference can fail after aggressive updates to $\alpha$ in the M-Step, due to the approximate nature of the log-likelihood evaluation with the linearization assumption.
A KL bound, simplified to a bound $\delta_\alpha$ on the update ratio, smooths the optimization by limiting aggressive updates.
For tuning, $\delta_\alpha$ should smooth out any large updates to $\alpha$ while limiting the impact to the rate of convergence.  
Note that as $\alpha$ is increasing, it is numerically easier to work with $\alpha\inv$, which tends to zero, so the limiting is implemented in practice as $\delta_{\alpha\inv}$ acting on $\nicefrac{\alpha^i}{\alpha^{i+1}}$.
Figure \ref{fig:sensitivity} demonstrates the behaviour of the hyperparameters for the Cartpole swing-up task. 
\section{Experimental Details}
\subsection{Equivalence with finite-horizon LQR by Dynamic Programming}
\label{subsec:lqr_experiment_details}
 \begin{align}
     \mathbf{x}_{t+1} &=\left[\begin{array}{cc} 1.1 & 0 \\ 0.1 & 1.1 \end{array} \right]\mathbf{x}_t + \left[\begin{array}{c} 0.1 \\ 0 \end{array} \right]\mathbf{u}_t + \left[\begin{array}{c} -1 \\ -2 \end{array} \right]\label{eq:lin_sys}\\
     \mQ &= \left[\begin{array}{cc} 10 & 0 \\ 0 & 10 \end{array} \right],\;\;
     \mR = [1],\;\;
     \vx_g = \left[\begin{array}{c} 10 \\ 10 \end{array} \right],\;\;
     \vu_g = [0],\;\;
     \alpha = 1\mathrm{e}{5},\;\;
     \sigr{u}{}{t} = [100]
 \end{align}
 \newcommand{\sne}[1]{1\mathrm{e}{\text{-}#1}}
 \subsection{Evaluation on nonlinear trajectory optimization tasks}
 \label{subsec:nonlinear_eval}
 Both iLQR and GPS required the cost function in Table \ref{tab:nonlinear_param} to be scaled in order to have good numerics. In Table \ref{tab:nonlinear_ilqr_param} and \ref{tab:nonlinear_gps_param} we refer to this has $\alpha$ (as it performs the same role as the \itwoc{} parameter). 
 Additionally, iLQR and GPS were enable to optimize without a random initialization.
 In order to compare with \itwoc{}, which initializes by design with fixed priors, the random initialisation was set to have a smallest amplitude that allowed optimization to take place.
 All algorithms achieved faster converged with random initialisation, however such `warm start' strategies were not the focus of this work, instead we wished to focus on \itwoc{} strength in deterministic initialisation. For these experiments we use the terminal cost $\mQ_f = \mQ$.
 \label{subsec:nonlinear_experiment_details}
 {\renewcommand{\arraystretch}{1}
 	\begin{table}[t]
 		\centering
 		\begin{tabular}{p{1.6cm}p{3cm}p{1.8cm}p{2.7cm}p{0.6cm}l}
 			\toprule
 			Environment & $\vz$ & $\vz_g$ & $\bm{\Theta}=\text{diag}(\mQ,\mR)$ & $\vu_{\text{limit}}$ & $\sigEta{~}$\\
 			\midrule
 			Pendulum & $[\sin\theta,\cos\theta,\dot{\theta}, u]\tran$ & $[0,1,0,0]\tran$& $\text{diag}(1,100,1,1)$ & $[-2,2]$ & $\text{diag}(\eps_1,\eps_3)$\\
 			\midrule
 			Cartpole & $[x,\sin\theta,\cos\theta,\dot{x},\dot{\theta}, u]\tran$ & $[0,0,1,0,0,0]\tran$ & $\text{diag}(1,1,100,1,1,1)$ & $[-5,5]$ & $\text{diag}(\eps_1,\eps_1,\eps_2,\eps_2)$\\
 			\midrule
 			\begin{minipage}{5mm} 
 			Double\\Cartpole \end{minipage} & 
 			\begin{minipage}{5mm} 
 				$[x,\sin\theta_1,\cos\theta_1,\sin\theta_2,\\
 				~\;\cos\theta_2,\dot{x},\dot{\theta_1},\dot{\theta_2}, u]\tran$ \end{minipage} 
 			  & 
 			  \begin{minipage}{5mm} 
 			  $[0,0,1,0,1,\\~\;0,0,0,0]\tran$ 
 			  \end{minipage} 
 			  & \begin{minipage}{5mm} 
 			  	$\text{diag}(1,1,100,1,100,\\~\hspace{7.5mm}1,1,1,1)$ 
 			  \end{minipage}
 			  & \begin{minipage}{5mm} $[-10,\\~\hspace{2mm}10]$
 				\end{minipage} 
 			  & \begin{minipage}{5mm} 
 			  	$\text{diag}(\eps_1,\eps_1, \eps_1,\\~\hspace{7.5mm}\eps_2,\eps_2,\eps_2)$ 
 			  \end{minipage}\\
 			\bottomrule\vspace{2mm}
 		\end{tabular}
 		\caption{Environment parameters of the nonlinear tasks. $\eps_1\!=\!\sne{12}$, $\eps_2\!=\!\sne{6}$ and $\eps_3\!=\!\sne{3}$.}
 		\label{tab:nonlinear_param}
 	\end{table}
 }
\newpage
{\renewcommand{\arraystretch}{1}
	\begin{table}[H]
		\centering
		\begin{tabular}{lllll}
			\toprule
			Environment & $\sigr{u}{}{}$ (init.) & $\alpha$ (init.) & $\delta_{\alpha\inv}$ & \\
			\midrule
			Pendulum & $[0.2]$  & $1/100$  & $0.99$  & \\
			Cartpole & $[0.25]$ & $1/67$   & $0.993$ &\\
			Double Cartpole & $[0.04]$ & $1/90$ & $0.9995$ & \\
			\bottomrule\vspace{2mm}
		\end{tabular}
		\caption{\itwoc{} parameters for the nonlinear trajectory optimization tasks.}
		\label{tab:nonlinear_i2c_param}
	\end{table}
}
{\renewcommand{\arraystretch}{1}
	\begin{table}[H]
		\centering
		\begin{tabular}{lllll}
			\toprule
			Environment & $\lambda$ range & $\lambda_{\text{multiplier}}$ & $\sigma_k$ (init.)  & $\alpha$\\
			\midrule
			Pendulum & $[1-\sne{9}]$ & $1.002$ & $\sne{2}$ & $\sne{4}$ \\
			Cartpole & $[1-\sne{7}]$ & $1.001$ & $\sne{2}$ & $\sne{3}$\\
			Double Cartpole & $[1-\sne{7}]$ & $1.001$  & $\sne{2}$ & $\sne{3}$ \\
			\bottomrule\vspace{2mm}
		\end{tabular}
		\caption{iLQR parameters for the nonlinear trajectory optimization tasks.}
		\label{tab:nonlinear_ilqr_param}
	\end{table}
}

{\renewcommand{\arraystretch}{1}
	\begin{table}[H]
		\centering
		\begin{tabular}{lllll}
			\toprule
			Environment & $\mSigma_{\text{Explore}}$  & KL bound & $\sigma_k$ (init.) & $\alpha$\\
			\midrule
			Pendulum & $[2.0]$ & $0.07$  & $\sne{2}$ & $\sne{4}$\\
			Cartpole & $[1.25]$& $1.0$   & $\sne{1}$ & $\sne{3}$ \\
			Double Cartpole & $[5.0]$ & $0.75$ & $\sne{1}$ & $\sne{3}$ \\
			\bottomrule\vspace{2mm}
		\end{tabular}
		\caption{GPS parameters for the nonlinear trajectory optimization tasks}
		\label{tab:nonlinear_gps_param}
	\end{table}
}
\begin{figure}[b!]
	\begin{minipage}[t]{\columnwidth}
		\centering
		\begin{tikzpicture}

\definecolor{color0}{rgb}{0.12156862745098,0.466666666666667,0.705882352941177}
\definecolor{color1}{rgb}{1,0.498039215686275,0.0549019607843137}
\definecolor{color2}{rgb}{0.172549019607843,0.627450980392157,0.172549019607843}

\begin{axis}[
hide axis,
width=14cm,
xmin=10, xmax=50,
ymin=0, ymax=1.0,
legend cell align={center},
legend columns=3,
legend style={/tikz/every even column/.append style={column sep=1.5cm}}]
]

\addlegendimage{no markers, line width=1.0pt, color0}
\addlegendentry{\small I2C};
\addlegendimage{no markers, line width=1.0pt, color1}
\addlegendentry{\small iLQR };
\addlegendimage{no markers, line width=1.0pt, color2}
\addlegendentry{\small GPS};

\end{axis}

\end{tikzpicture}
		\vspace{2mm}
	\end{minipage}
	\hfill
	\begin{minipage}[b]{\columnwidth}
		\centering
		\input{fig/Cartpole_traj.tex}
		\caption{Comparison of the state-action trajectories of \itwoc{}, iLQR and GPS on the Cartpole swing-up task after convergence.}
		\label{fig:cartpole_traj}
	\end{minipage}
\end{figure}
\begin{figure}[t!]
	\begin{minipage}[t]{\columnwidth}
		\centering
		\begin{tikzpicture}

\definecolor{color0}{rgb}{0.12156862745098,0.466666666666667,0.705882352941177}
\definecolor{color1}{rgb}{1,0.498039215686275,0.0549019607843137}
\definecolor{color2}{rgb}{0.172549019607843,0.627450980392157,0.172549019607843}

\begin{axis}[
hide axis,
width=14cm,
xmin=10, xmax=50,
ymin=0, ymax=1.0,
legend cell align={center},
legend columns=3,
legend style={/tikz/every even column/.append style={column sep=1.5cm}}]
]

\addlegendimage{no markers, line width=1.0pt, color0}
\addlegendentry{\small I2C};
\addlegendimage{no markers, line width=1.0pt, color1}
\addlegendentry{\small iLQR };
\addlegendimage{no markers, line width=1.0pt, color2}
\addlegendentry{\small GPS};

\end{axis}

\end{tikzpicture}
	\end{minipage}
	\hfill
	\begin{minipage}[b]{\columnwidth}
		\begin{center}
		\input{fig/DoubleCartpole_traj.tex}
		\caption{Comparison of the state-action trajectories of \itwoc{}, iLQR and GPS on the Double Cartpole swing-up task after convergence.}
		\label{fig:double_cartpole_traj}
	    \end{center}
	\end{minipage}
\end{figure}

\flushbottom

\end{document}